\documentclass[11pt]{article}

\usepackage[final]{acl}

\usepackage{times}
\usepackage{latexsym}

\usepackage[T1]{fontenc}
\usepackage[utf8]{inputenc}

\usepackage[expansion=true,stretch=50,shrink=50,step=1,protrusion=true]{microtype}
\usepackage{inconsolata}

\usepackage{graphicx}
\usepackage{booktabs} %

\usepackage{amsmath}
\usepackage{amssymb}

\usepackage{url}

\usepackage{enumitem}
\usepackage{multirow}

\DeclareMathAlphabet{\mathmybb}{U}{bbold}{m}{n}
\newcommand{\one}{\mathmybb{1}}

\title{VTBench: Evaluating Visual Tokenizers for Autoregressive Image Generation}

\author{
 \textbf{Huawei Lin\textsuperscript{1}},
 \textbf{Tony Geng\textsuperscript{2}},
 \textbf{Zhaozhuo Xu\textsuperscript{3}},
 \textbf{Weijie Zhao\textsuperscript{1}}
\\
\\
 \textsuperscript{1}Rochester Institute of Technology,
 \textsuperscript{2}Rice University,\\
 \textsuperscript{3}Stevens Institute of Technology
}

\begin{document}
\maketitle

\begin{abstract}
Autoregressive (AR) models have recently shown strong performance in image generation, where a key component is the visual tokenizer (VT) that maps continuous pixels into discrete tokens. The reconstruction fidelity of a VT plays an important role in AR image generation quality. Existing discrete VTs often lose fine details and text during reconstruction, contributing to degraded generation quality. Current benchmarks primarily evaluate end-to-end generation quality, making it difficult to isolate VT performance. We introduce \textbf{VTBench}, a comprehensive benchmark for systematic evaluation of visual tokenizers across three core tasks: image reconstruction, detail preservation, and text preservation. Continuous VAEs show a clear overall advantage over discrete VTs in reconstruction fidelity, detail preservation, and text accuracy. Discrete VTs often produce distorted reconstructions and fail to preserve fine textures and text. We release VTBench publicly to facilitate future research on robust, general-purpose visual tokenizers.\footnote{\sbox{0}{Dataset:\ }\begin{tabular}[t]{@{}l@{\ }p{\dimexpr\linewidth-1.8em-\wd0\relax}@{}}Code: & \url{https://github.com/huawei-lin/VTBench}.\\Dataset: & \url{https://huggingface.co/datasets/huaweilin/VTBench}.\end{tabular}}
\end{abstract}

\begin{figure*}[t]
  \centering
  \includegraphics[width=\textwidth]{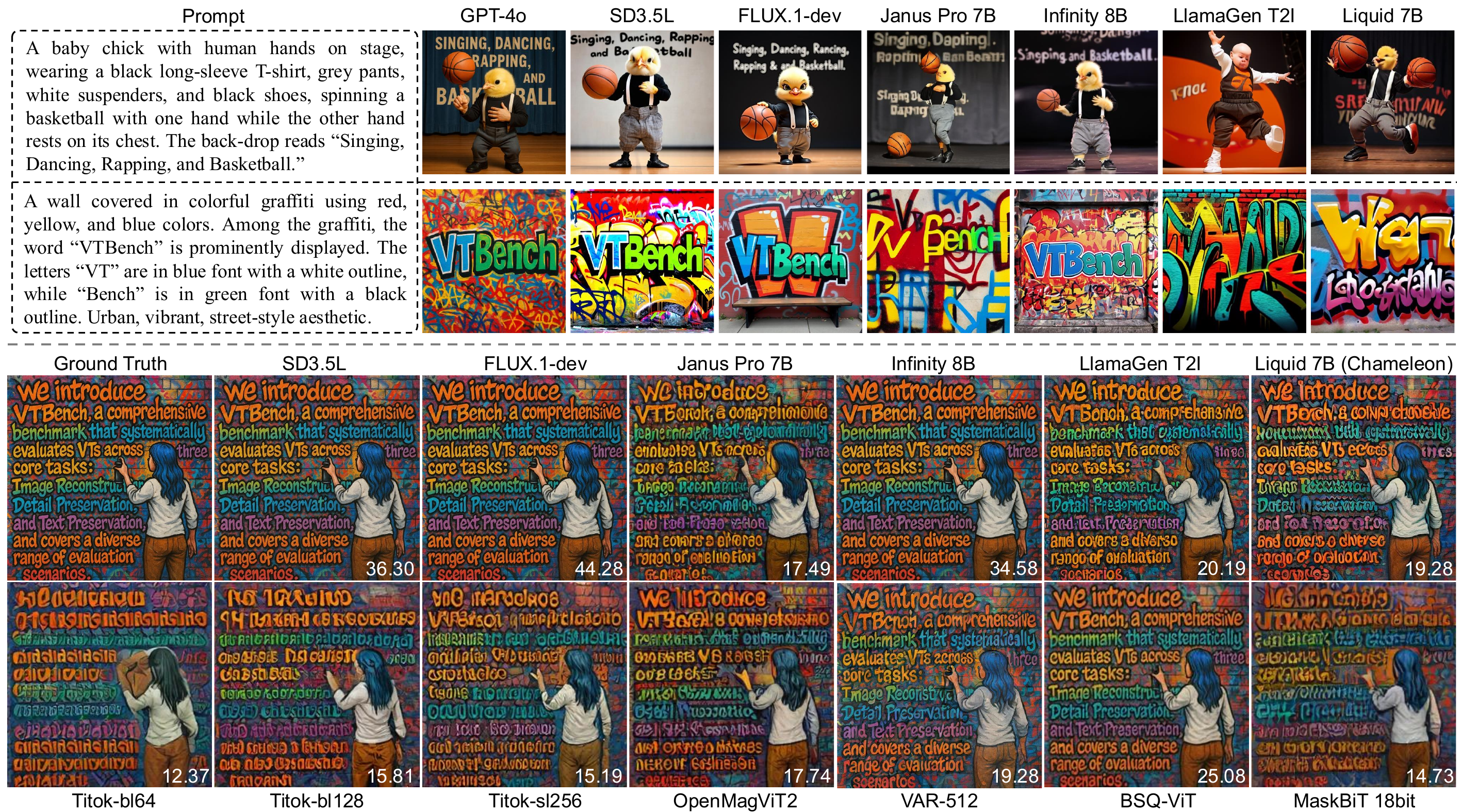}
  \caption{Image generation and reconstruction across different models. \textbf{(Top)} Images generated from prompts using various models. \textbf{(Bottom)} Reconstructions of the ground truth image using VTs from different models. PSNR $\uparrow$ values (shown in white) indicates reconstruction fidelity.}
  \label{fig:comparison_of_generation}
\end{figure*}

\section{Introduction}

Large Language Models (LLMs) have demonstrated impressive generalization across a wide range of tasks, including reasoning~\citep{DBLP:conf/nips/NingWL0Y0BDY024, DBLP:journals/corr/abs-2407-11511}, question answering~\citep{DBLP:conf/emnlp/0010LIWYBB24, DBLP:conf/eccv/SimaRCCZXBLGL24} and text generation~\citep{DBLP:journals/corr/abs-2302-13971, DBLP:journals/corr/abs-2412-19437, DBLP:journals/corr/abs-2401-04088, DBLP:conf/acl/LinLX024}. Recent advances suggest that integrating visual understanding and generation into the LLM framework could lead to unified, general-purpose multimodal models~\citep{fluid, Unified-IO2, Chameleon, show-o, seed-x}.

\begin{figure*}[t!]
  \centering
  \includegraphics[width=\textwidth]{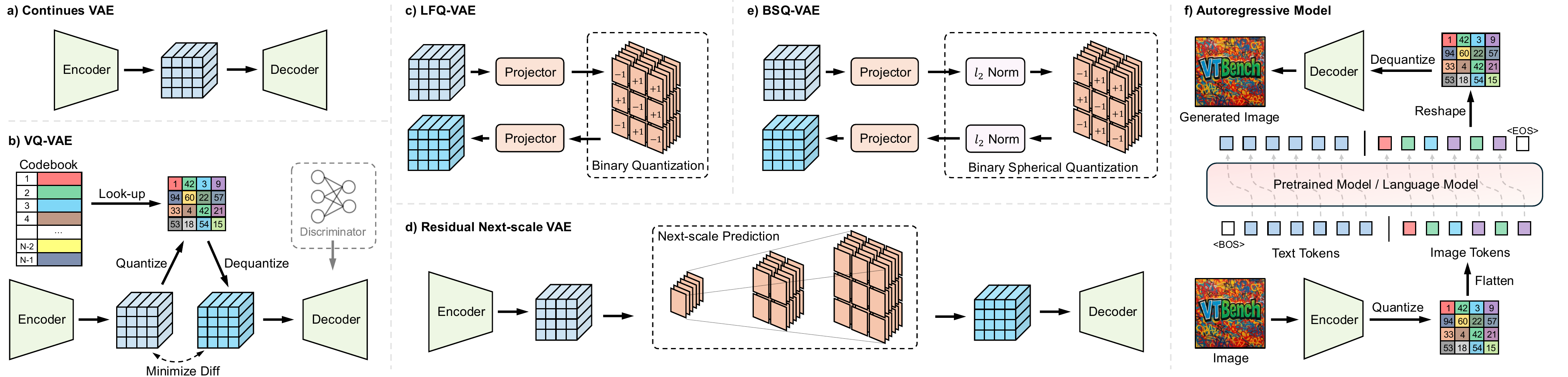}
  \caption{Overview of visual tokenizer architectures and integration with AR image generation.}
  \label{fig:VT_framework}
\end{figure*}

\looseness=-1
\textbf{Visual Tokenizer.} In diffusion-based image generation, images are typically compressed into a continuous latent space using a variational autoencoder (VAE), allowing the model to operate in a lower-dimensional but still continuous domain~\citep{DBLP:conf/nips/HoJA20, DBLP:journals/corr/abs-2112-10752}. However, when integrating visual understanding and generation into LLMs, images must be converted into discrete token sequences, similar to word or subword tokens in natural language processing~\citep{vqgan, DBLP:conf/cvpr/ChangZJLF22, DBLP:journals/corr/abs-2305-12223}. Figure~\ref{fig:VT_framework} illustrates various visual tokenizer architectures and the autoregressive modeling used for image generation.

\textbf{Vector Quantized Tokenization.} To enable such tokenization of visual inputs, Vector Quantized Variational Autoencoder (VQ-VAE) is proposed to encode images into a continuous latent space and then map each latent vector to the nearest entry in a learned codebook, producing a discrete index per spatial location~\citep{vqvae, vqgan}. However, scaling the codebook often leads to codebook collapse, where only a small portion of entries are used, reducing representational capacity. Additionally, nearest-neighbor search introduces computational inefficiency during generation~\citep{LFQ, infinity}. To address these limitations, Lookup-Free Quantization (LFQ) eliminates the embedding lookup by projecting continuous features into a binary latent space, enabling discrete tokenization without explicit search~\citep{LFQ}. To constrain error and enhance stability, Binary Spherical Quantization (BSQ) introduces $\ell_2$ normalization, smoothing the representation~\citep{BSQViT}.

\textbf{Next-Scale Tokenization.} However, AR modeling with existing visual tokenizers often violates the unidirectional dependency assumption and disrupts spatial locality due to the flattening of 2D token grids. To overcome these limitations, \citet{VAR} proposed Visual Autoregressive Modeling (VAR), which encodes images into multiple levels of tokens by progressively quantizing the residual information at each scale. This hierarchical tokenization preserves spatial structure and allows the model to capture fine-grained details in a coarse-to-fine manner. In Figure~\ref{fig:VT_framework}, we refer to this architecture as ``Residual Next-Scale VAE''. Building on this foundation, Infinity enhances VT by integrating BSQ, enabling extremely large vocabularies and efficient scaling. This allows Infinity to generate high-resolution images with unprecedented detail~\citep{infinity}.

Despite the growing number of AR models for image generation~\citep{DBLP:conf/nips/LiTLDH24, DBLP:journals/pami/BieYZGZYWHGCHTS25, DBLP:journals/corr/abs-2409-04429}, most existing methods are still limited to relatively simple datasets such as ImageNet and fall significantly behind  
diffusion models in terms of generation quality and detail preservation. As shown in Figure~\ref{fig:comparison_of_generation}(Top), which presents outputs from several SOTA text-to-image models prompted with highly detailed descriptions, only the images generated by diffusion models (SD3.5L~\citep{stable-diffusion} and FLUX.1-dev~\citep{flux2024}) and GPT-4o (architecture currently undisclosed~\citep{yan2025gpt, li2025have}) exhibit high-quality synthesis. We hypothesize that the visual tokenizer plays a critical role in this quality gap. To investigate this, we conduct a reconstruction experiment using the images generated by GPT-4o as the ground truth. As shown in Figure~\ref{fig:comparison_of_generation}(Bottom), we reconstruct the ground truth image using different VTs from various AR models. The results reveal substantial information loss, including blurred or unreadable text, loss of fine-grained visual details, and noticeable structural distortions. These failures indicate that current VTs struggle to generate accurate and expressive latent representations, which limits the overall image generation quality of AR models.

\textbf{\textit{Why Benchmarking VT Matters?}} The reconstruction failures observed in Figure~\ref{fig:comparison_of_generation} highlight a fundamental issue: current visual tokenizers often fail to preserve fine-grained details and semantic integrity during the quantization process~\citep{LFQ, BSQViT}. Poor reconstruction fidelity reveals losses in the visual representations used by AR models, making visual tokenization an important bottleneck in downstream image generation. Despite their central role, existing evaluation protocols focus almost exclusively on end-to-end generation quality, without isolating the contribution or limitations of the VT itself. This leaves several critical issues unaddressed:
(1) \textbf{Lack of VT-Specific Evaluation:} VT reconstruction quality is an important factor in downstream AR image generation, because it reflects how faithfully visual information is preserved~\citep{LFQ}. Yet, most VTs are only evaluated on limited datasets like ImageNet~\citep{imagenet}, and there is a lack of dedicated benchmarks designed specifically to measure VT effectiveness across diverse scenarios.
(2) \textbf{Benchmark Misalignment:} Existing benchmarks evaluate overall image generation (e.g., ImageNet~\citep{imagenet}, GenEval~\citep{DBLP:conf/nips/GhoshHS23}, T2i-compbench~\citep{DBLP:conf/nips/HuangSXLL23, DBLP:journals/pami/HuangDSXLL25}), rather than isolating the visual tokenizer contribution, making it difficult to diagnose or improve this key component.
(3) \textbf{Inadequate Evaluation Metrics:} Commonly used metrics such as FID are insufficient to capture fine-grained failures like high-frequency detail loss or incorrect text reconstruction, which are essential for many multimodal and LLM-centric tasks.

To address these issues, in this paper, we introduce VTBench, a comprehensive benchmark covering major VT families and three intrinsic capabilities: reconstruction fidelity, detail preservation, and text preservation. VTBench provides a systematic framework for understanding and improving VTs as standalone components in AR image generation pipelines. Through extensive experiments on a wide range of VTs used in SOTA AR models, we uncover the following key findings:
(1) Continuous VAEs show a clear overall advantage over discrete VTs in reconstruction fidelity, detail preservation, and text accuracy.
(2) The evaluated discrete VTs exhibit resolution-dependent failures or input-size constraints under our protocol; successful reconstruction at a high square resolution does not imply arbitrary-resolution robustness.
(3) Many AR models (e.g., Chameleon~\citep{Chameleon}, Liquid~\citep{liquid}, Anole~\citep{anole}) reuse the same open-source VT, yet there is currently no strong, general-purpose VT available for reuse, highlighting the lack of a reliable, high-quality open-source solution in this space.

\textbf{Contributions.} The main contributions of this paper are:
\vspace{-.05in}
\begin{itemize}[itemsep=2pt, topsep=0pt, leftmargin=10pt, parsep=0pt, partopsep=0pt]
\item We introduce VTBench, a high-quality and comprehensive benchmark designed specifically for evaluating VTs in the context of AR image generation.
\item We design three tasks: Image Reconstruction, Detail Preservation, and Text Preservation, collectively providing a multi-faceted framework for assessing visual tokenizers. These tasks cover diverse evaluation aspects, including high-resolution inputs, multilingual text scenarios (Chinese, Korean, Japanese, Hindi), and varying-resolution conditions.
\item We conduct extensive experiments on discrete VTs and continuous VAEs, with GPT-4o as a system-level reference. Our evaluation is designed to assess both image quality and text preservation through a diverse set of quantitative metrics.
\item We open-source both the codebase and the VTBench dataset to foster further research in visual tokenization for autoregressive image generation. Our codebase is designed to be lightweight and easy to run, requiring minimal setup and no complex configuration.
\end{itemize}

\section{Background}

In this section, we introduce image generation methods, focusing on both diffusion and AR models. We then discuss the role of visual tokenizers in AR pipelines, reviewing several architectures including VQ-VAE, LFQ-VAE, BSQ-VAE, and Residual Next-Scale VAE, as illustrated in Figure~\ref{fig:VT_framework}. Finally, we highlight recent advances in image generation with GPT-4o, which motivate the need for deeper evaluation of visual tokenization quality.

\subsection{Diffusion vs. Autoregressive Models}
Modern image generation methods are mainly dominated by two families: diffusion models~\citep{DBLP:journals/pami/CroitoruHIS23, DBLP:journals/csur/YangZSHXZZCY24} and AR models~\citep{DBLP:journals/corr/abs-2503-13436, janus-pro, titok}. Diffusion models learn to iteratively denoise a sample from pure noise to a realistic image using a learned reverse diffusion process, such as Stable Diffusion~\citep{stable-diffusion}, which have demonstrated SOTA performance in high-quality image synthesis. In contrast, AR models decompose image generation into a sequence modeling problem, where an image is represented as a sequence of discrete tokens predicted one at a time~\citep{ma2025token, openmagvit2, DBLP:conf/cvpr/YuCSL0CH0HE023}, as shown in Figure~\ref{fig:VT_framework}(f). AR models benefit from the scaling properties of LLMs and enable seamless integration with multimodal pipelines. However, AR image generation relies heavily on high-quality visual tokenization to convert pixel data into token sequences and reconstruct images from these tokens, making the design of effective visual tokenizers a central challenge in AR-based image generation.

\subsection{Visual Tokenizers}

To enable AR models to process image inputs, images must be converted into discrete token sequences. This is achieved via visual tokenizers that compress high-dimensional image data into compact, symbolic representations. We briefly introduce representative tokenization approaches here; a detailed version with mathematical formulations is included in Appendix~\ref{apd:vt-background}.

\begin{figure*}[t]
  \centering
  \includegraphics[width=\textwidth]{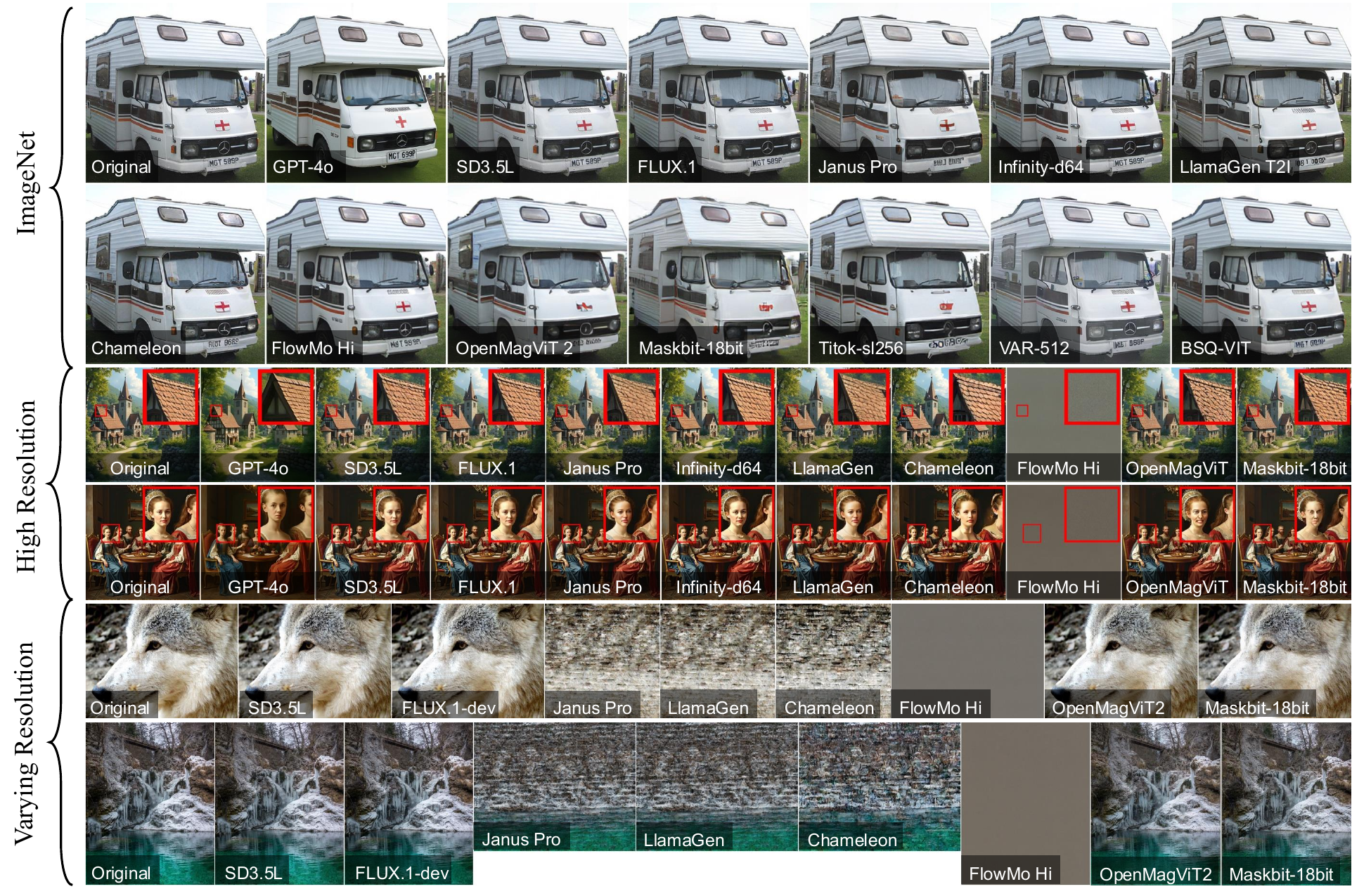}
  \caption{Examples of task 1: (1) ImageNet, (2) High Resolution and (3) Varying Resolution. FlowMo Hi produces corrupted images under our direct whole-image High Resolution and Varying Resolution protocol, while Janus Pro, LlamaGen, and Chameleon generate images with incorrect resolution and distorted semantics.}
  \label{fig:task1_examples}
\end{figure*}

\begin{itemize}[itemsep=2pt, topsep=0pt, leftmargin=10pt, parsep=0pt, partopsep=0pt]
\vspace{-.05in}
\item \textbf{Continuous VAE.} In diffusion models, a common approach is to use a continuous Variational Autoencoder (VAE)~\citep{stable-diffusion, DBLP:conf/iccv/PeeblesX23} as a feature compressor as shown in Figure~\ref{fig:VT_framework}(a). These VAEs encode the input image into a continuous latent space, typically using convolutional encoders and decoders, enabling high-quality image reconstructions.
\item  \textbf{VQ-VAE.} Vector Quantized Variational Autoencoder (VQ-VAE)~\citep{vqvae} is one of the earliest and most widely adopted discrete visual tokenizers. It encodes an image into a latent feature map and then quantizes each spatial location to the closest vector in a learned codebook. The resulting discrete indices form a token grid that can be used in AR models.
\item \textbf{LFQ-VAE.} Lookup-Free Quantization (LFQ)~\citep{LFQ} eliminates the nearest-neighbor lookup and codebook by projecting features into a binary latent space by a learnable mapping.
\item \textbf{BSQ-VAE.} Binary Spherical Quantization (BSQ)~\citep{BSQViT} extends LFQ by applying $\ell_2$ normalization to the latent features before quantization. This constrains the representation to a unit hypersphere, effectively reducing quantization error and leading to smoother latent spaces. BSQ allows for finer-grained tokenization while maintaining efficient binary encoding, which is beneficial for downstream tasks sensitive to visual detail.
\item \textbf{Residual Next-Scale VAE.} applies hierarchical tokenization by encoding images in a coarse-to-fine manner. It quantizes residual information across multiple spatial scales, enabling preservation of structure and fine details, especially in high-resolution settings~\citep{VAR}.
\end{itemize}

\subsection{GPT-4o Image Generation}
\looseness=-1
OpenAI recently introduced new image generation and editing capabilities in GPT-4o~\citep{openai2025image}, showcasing remarkable performance in both tasks. Although the precise architecture and workflow of GPT-4o remain undisclosed, recent studies suggest that the model may employ an autoregressive backbone in conjunction with a diffusion-based decoder for image synthesis\citep{yan2025gpt, li2025have}. In this paper, we present extensive experiments on GPT-4o to investigate its architecture and workflow, with a focus on how the visual tokenizer influences image quality.

\section{VTBench: Task \& Evaluation Metrics}
To systematically evaluate VTs in AR image generation pipelines, we propose VTBench, a comprehensive benchmark that isolates and diagnoses the capabilities and limitations of VTs across three critical tasks: (1) Image Reconstruction, (2) Detail Preservation, and (3) Text Preservation. Data valuation and influence estimation help assess how training data affect model behavior~\citep{DBLP:conf/naacl/PanLRCYZZX25, lin2026dmin}. Each task is designed to stress different aspects of tokenization quality, using diverse data conditions.

\begin{figure*}[t]
  \centering
  \includegraphics[width=.9\textwidth]{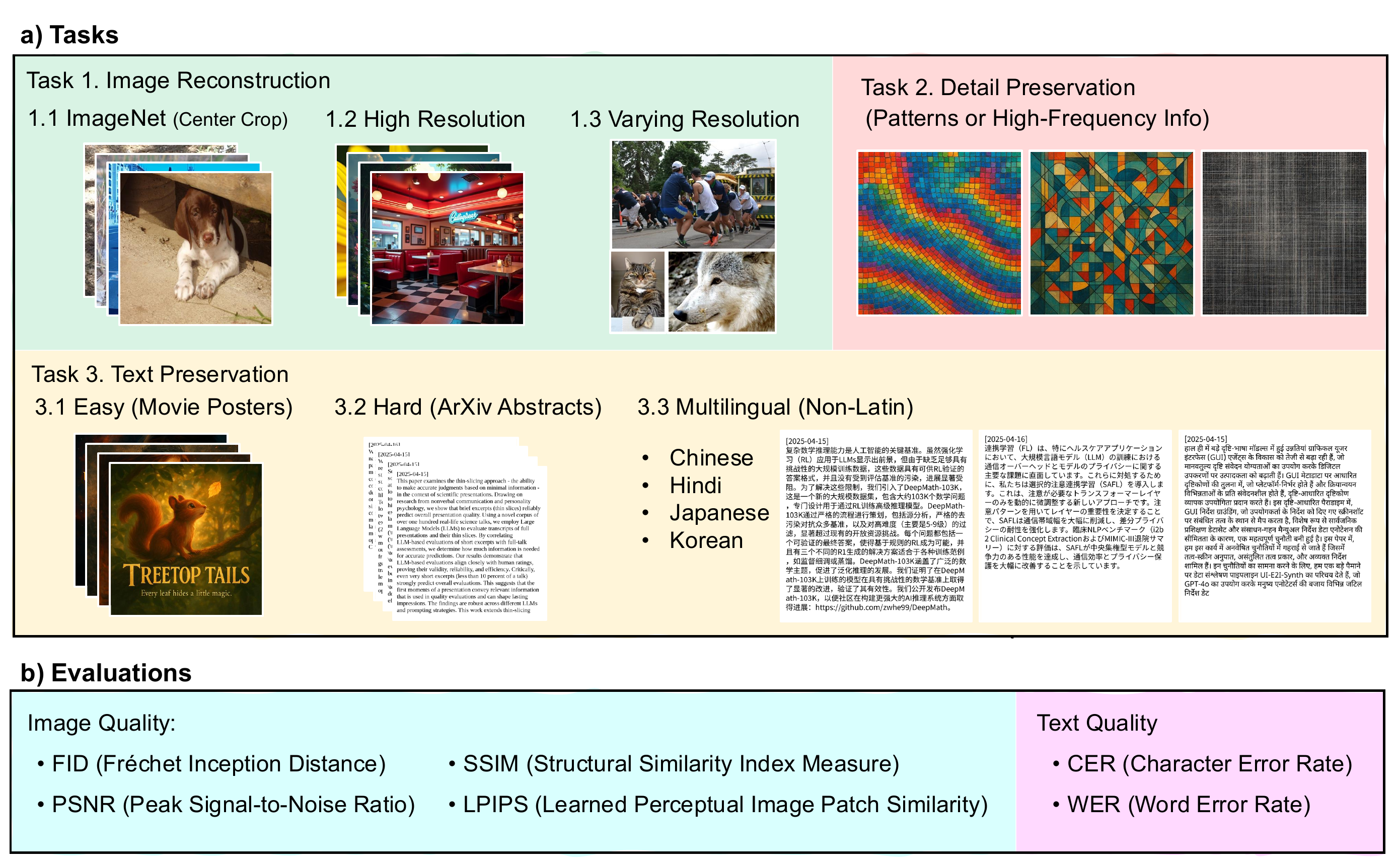}
  \caption{Overview of VTBench construction. (a) VTBench consists of three core tasks for evaluating visual tokenizers. (b) Evaluation include both image quality metrics and text-specific metrics.}
  \label{fig:VTBench}
\end{figure*}

\subsection{Task 1: Image Reconstruction}
This task evaluates the fundamental ability of a VT to reconstruct an image from its tokenized representation. As the interface between high-dimensional pixel inputs and the discrete token sequences consumed by AR models, the VT's reconstruction fidelity is an important factor in downstream image generation quality. If the tokenizer fails to preserve fine details or text, the resulting representation may limit the fidelity of generated images. Reconstruction quality therefore diagnoses a generation-relevant bottleneck.

To isolate the intrinsic capacity of each VT, we assess image reconstruction performance independent of generation. Specifically, we evaluate three settings designed to test robustness across resolution and scale: (1) ImageNet (model-specific input size), (2) High Resolution ($1024\times 1024$ inputs), and (3) Varying Resolution (images with diverse, unconstrained dimensions), as illustrated in Figure~\ref{fig:VTBench}(a).

\textbf{ImageNet.} We begin with the validation subset of the standard ImageNet-1k dataset~\citep{imagenet}, containing 50,000 images. This benchmark serves as a canonical reference point, as most open-source VTs are evaluated on it. The task assesses whether the VT preserves general semantics and structural integrity in a typical low-resolution setting. In this subtask, all images are center-cropped and resized to model-specific input sizes as listed in Table~\ref{tbl:task1}.

\textbf{High Resolution.} To evaluate scalability, we collect 100 high-resolution images ($1024 \times 1024$), and assess the quality of their reconstructions. This setting exposes limitations in the VT's ability to preserve visual fidelity across larger spatial extents, where coarse quantization or resolution mismatches often degrade reconstruction quality.

\textbf{Varying Resolution.} In real-world scenarios, image resolutions vary widely, posing challenges for models that assume fixed-size inputs. To evaluate the robustness of VTs under such conditions, we use a mixed-resolution dataset, specifically the test subset of DIV2K~\citep{Agustsson_2017_CVPR_Workshops}, which includes 100 high-quality images with a broad range of dimensions. Unlike continuous VAEs, most discrete tokenizers are not resolution-agnostic and require fixed-size inputs. This task quantifies how such constraints affect reconstruction accuracy and the model's flexibility in handling diverse input sizes, with a detailed architectural analysis presented in Appendix~\ref{apd:resolution_limitations}.

\subsection{Task 2: Detail Preservation}
While overall reconstruction quality is important, many downstream tasks, such as object recognition, editing, and captioning, depend on the preservation of fine-grained visual details. These include textures, facial features, edges, and small objects that may occupy only a few pixels. Such details are often the first to be degraded or lost during quantization, especially when the codebook or token representation lacks sufficient expressiveness. A VT that fails to preserve these features will fundamentally limit the quality and realism of generated images. Therefore, this task focuses on measuring how well VTs retain high-frequency information crucial for perceptual fidelity. We collect 100 images rich in detailed, high-frequency semantic content. These 100 images form a targeted stress test of detail preservation. We provide a more detailed definition of ``visual details'' and quantify that images in Task 2 contain more fine-grained information than other tasks in Appendix~\ref{apd:detail_definition}.

\begin{table*}[t]
\centering
\caption{Evaluation of VTs across three reconstruction settings in Task 1.}\label{tbl:task1}
\resizebox{.9\textwidth}{!}{%
\begin{tabular}{lcc|ccccc|ccccc|cccc}
\toprule
\midrule
\multirow{2}{*}{Method} & \multirow{2}{*}{Params} & \multirow{2}{*}{Type} & \multicolumn{5}{c|}{ImageNet}                                                                  & \multicolumn{5}{c|}{High Resolution}                                                           & \multicolumn{4}{c}{Varying Resolution}                                   \\
                        &                         &                       & Resolution         & PSNR$\uparrow$   & SSIM$\uparrow$  & LPIPS$\downarrow$ & FID$\downarrow$ & Resolution         & PSNR$\uparrow$   & SSIM$\uparrow$  & LPIPS$\downarrow$ & FID$\downarrow$ & PSNR$\uparrow$   & SSIM$\uparrow$  & LPIPS$\downarrow$ & FID$\downarrow$ \\\midrule
FlowMo Lo               & 945M                    & LFQ                   & $256 \times 256$   & 20.2232          & 0.5878          & 0.1031            & 0.8227          & $1024 \times 1024$ & 11.3631          & 0.3644          & 0.6913            & 465.3528        & 11.4990          & 0.2854          & 0.7379            & 442.0812        \\
FlowMo Hi               & 946M                    & LFQ                   & $256 \times 256$   & 22.4923          & 0.7103          & 0.0684            & 0.4834          & $1024 \times 1024$ & 11.3077          & 0.3874          & 0.6937            & 439.7494        & 11.4571          & 0.2862          & 0.7353            & 412.3167        \\
MaskBiT 16bit           & 54M                     & LFQ                   & $256 \times 256$   & 19.6448          & 0.5259          & 0.1246            & 1.1473          & $1024 \times 1024$ & 28.6892          & 0.9074          & 0.0313            & 24.2461         & 23.9538          & 0.8037          & 0.0405            & 22.4535         \\
MaskBiT 18bit           & 55M                     & LFQ                   & $256 \times 256$   & 19.7695          & 0.5352          & 0.1197            & 1.0334          & $1024 \times 1024$ & 28.9353          & 0.9135          & 0.0295            & 23.3024         & 23.9289          & 0.8070          & 0.0394            & 23.3959         \\
Titok-l32               & 641M                    & VQ                    & $256 \times 256$   & 15.0997          & 0.3533          & 0.2922            & 1.8181          & $1024 \times 1024$ & -                & -               & -                 & -               & -                & -               & -                 & -               \\
Titok-b64               & 204M                    & VQ                    & $256 \times 256$   & 16.1520          & 0.3903          & 0.2285            & 1.3542          & $1024 \times 1024$ & -                & -               & -                 & -               & -                & -               & -                 & -               \\
Titok-s128              & 83M                     & VQ                    & $256 \times 256$   & 16.7386          & 0.4123          & 0.1898            & 1.4186          & $1024 \times 1024$ & -                & -               & -                 & -               & -                & -               & -                 & -               \\
Titok-bl64              & 390M                    & VQ                    & $256 \times 256$   & 17.3930          & 0.4321          & 0.1926            & 1.7696          & $1024 \times 1024$ & -                & -               & -                 & -               & -                & -               & -                 & -               \\
Titok-bl128             & 390M                    & VQ                    & $256 \times 256$   & 18.4675          & 0.4877          & 0.1468            & 1.2328          & $1024 \times 1024$ & -                & -               & -                 & -               & -                & -               & -                 & -               \\
Titok-sl256             & 330M                    & VQ                    & $256 \times 256$   & 19.6375          & 0.5514          & 0.1108            & 0.8112          & $1024 \times 1024$ & -                & -               & -                 & -               & -                & -               & -                 & -               \\
OpenMagViT2             & 115M                    & LFQ                   & $256 \times 256$   & 20.0090          & 0.5786          & 0.1028            & 1.0598          & $1024 \times 1024$ & 30.7770          & 0.9344          & 0.0232            & 18.6649         & 24.2372          & 0.8133          & 0.0383            & 20.3403         \\
LlamaGen ds8            & 70M                     & VQ                    & $256 \times 256$   & 22.4021          & 0.6995          & 0.0592            & 0.4597          & $1024 \times 1024$ & 33.5435          & 0.9693          & 0.0112            & 10.1195         & 13.6772          & 0.2735          & 0.5401            & 330.8865        \\
BSQ-VIT                 & 174M                    & RBS                   & $256 \times 256$   & \textbf{25.8646} & \textbf{0.8331} & \textbf{0.0359}   & \textbf{0.4586} & $1024 \times 1024$ & -                & -               & -                 & -               & -                & -               & -                 & -               \\
VAR-256                 & 109M                    & RVQ                   & $256 \times 256$   & 20.3693          & 0.6035          & 0.0933            & 0.9080          & $1024 \times 1024$ & -                & -               & -                 & -               & -                & -               & -                 & -               \\\cline{1-8}
Janus Pro 1B/7B         & 72M                     & VQ                    & $384 \times 384$   & 22.4485          & 0.6793          & 0.0819            & 0.7111          & $1024 \times 1024$ & 29.0476          & 0.9208          & 0.0299            & 24.4655         & 13.1547          & 0.2342          & 0.5137            & 324.0125        \\\cline{1-8}
Chameleon               & 68M                     & VQ                    & $512 \times 512$   & 23.5837          & 0.7164          & 0.0672            & 0.8061          & $1024 \times 1024$ & 29.3743          & 0.9062          & 0.0279            & 22.6725         & 12.7700          & 0.2051          & 0.5206            & 331.3999        \\
LlamaGen ds16           & 72M                     & VQ                    & $512 \times 512$   & \textbf{24.2199} & 0.7568          & \textbf{0.0630}   & \textbf{0.5441} & $1024 \times 1024$ & 28.6424          & 0.9148          & 0.0328            & 25.0529         & 13.3258          & 0.2413          & 0.5171            & 327.8920        \\
LlamaGen ds16 T2I       & 72M                     & VQ                    & $512 \times 512$   & 24.0192          & 0.7493          & 0.0647            & 0.5528          & $1024 \times 1024$ & 29.0061          & 0.9206          & 0.0300            & 24.5831         & 13.1604          & 0.2344          & 0.5137            & 324.4282        \\
VAR-512                 & 109M                    & RVQ                   & $512 \times 512$   & 22.3828          & \textbf{0.7636} & 0.0727            & 0.7719          & $1024 \times 1024$ & -                & -               & -                 & -               & -                & -               & -                 & -               \\\cline{1-8}
Infinity-d32            & 110M                    & RBSQ                  & $1024 \times 1024$ & 33.4850          & 0.9582          & 0.0109            & 0.1002          & $1024 \times 1024$ & 35.5645          & 0.9732          & 0.0080            & 8.3090          & -                & -               & -                 & -               \\
Infinity-d64            & 110M                    & RBSQ                  & $1024 \times 1024$ & 36.0010          & 0.9766          & 0.0067            & 0.0511          & $1024 \times 1024$ & 37.5937          & 0.9823          & 0.0053            & 5.8734          & -                & -               & -                 & -               \\\midrule
SD3.5L                  & 83M                     & Continuous            & $1024 \times 1024$ & 38.8208          & 0.9836          & 0.0013            & 0.0121          & $1024 \times 1024$ & 38.4653          & 0.9839          & 0.0012            & 0.9787          & 30.6413          & 0.9507          & 0.0075            & 3.1084          \\
FLUX.1-dev              & 83M                     & Continuous            & $1024 \times 1024$ & \textbf{41.6134} & \textbf{0.9932} & \textbf{0.0006}   & \textbf{0.0050} & $1024 \times 1024$ & \textbf{41.4870} & \textbf{0.9921} & \textbf{0.0005}   & \textbf{0.4466} & \textbf{30.9502} & \textbf{0.9594} & \textbf{0.0068}   & \textbf{2.7123} \\
GPT-4o (ref.)                  & -                       & -                     & $1024 \times 1024$ & 19.3828          & 0.5769          & 0.1463            & 39.5056         & $1024 \times 1024$ & 15.3835          & 0.4642          & 0.2077            & 78.4113         & -          & -          & -            & -        
  \\
\midrule
\bottomrule
\end{tabular}%
}
\end{table*}

\subsection{Task 3: Text Preservation}

Text is a uniquely challenging and critical element in many real-world images, especially in domains such as documents, signage, user interfaces, and multimodal reasoning. Unlike general textures, text requires pixel-level precision, minor distortions can render characters unreadable, break words, or alter semantics entirely. Furthermore, text carries symbolic meaning that must be preserved for models to support language-image alignment, OCR, or instruction-following. Textual instructions also underpin reusable skills in LLM agents~\citep{DBLP:journals/corr/abs-2605-27366}. However, most VTs are not optimized for symbolic preservation, especially in multilingual or high-density settings. This task evaluates how VTs preserve textual content under varying complexity and linguistic diversity.

\textbf{Movie Posters (Easy).} In this subtask, we first craft descriptions for 100 fictional movies along with corresponding slogans, and then produce a poster for each movie. Each poster features a prominent title in large text, a slogan in smaller text, and a background that visually reflects the movie's theme or genre. 
The dataset spans a wide variety of movie types, providing diverse yet structured visual compositions. A strong VT should preserve both the overall image quality and the legibility of embedded text. Example posters are shown in Figure~\ref{fig:VTBench}.

\textbf{ArXiv Abstracts (Hard).} In addition to the movie posters, we introduce a more challenging subtask consisting of 100 images that render academic paper abstracts. These abstracts feature long sentences, dense layouts, varied font styles, and small font sizes, making this setting particularly demanding. The goal of this task is to evaluate the tokenizer's ability to preserve fine-grained textual content and maintain layout fidelity under complex visual conditions. To synthesize this dataset, we retrieved abstracts from papers published on April 16, 2025 to ensure that the rendered text images were not included in the training data of VT models.

\textbf{Multilingual (Non-Latin).}
To evaluate the cross-lingual robustness of visual tokenizers, we construct a multilingual benchmark consisting of non-Latin scripts, including Chinese, Hindi, Japanese, and Korean. Specifically, we translate the ArXiv abstract texts into each target language using GPT-4o, and then render the translated content into images following the same formatting and layout procedures as in the English version. For each language, we generate 100 text-rich images that reflect real-world typographic complexity. This subtask assesses whether VTs can preserve diverse character sets, linguistic structures, and writing systems that differ from Latin-based scripts.

\subsection{Evaluation Metrics}
To comprehensively evaluate VTs across the three core tasks of VTBench, we adopt a combination of standard and task-specific metrics to capture different dimensions of quality.

\textbf{Image Quality.}
To assess how well a VT reconstructs the original image from its tokenized representation, we employ four metrics: (1) \textbf{PSNR (Peak Signal-to-Noise Ratio):} Measures pixel-level fidelity between original and reconstructed images. Higher values indicate better reconstruction. (2) \textbf{SSIM (Structural Similarity Index):} Quantifies structural similarity by comparing luminance, contrast, and texture, and is more perceptually aligned than PSNR. (3) \textbf{LPIPS (Learned Perceptual Image Patch Similarity):} A perceptual similarity metric based on features from pretrained networks~\citep{DBLP:conf/cvpr/ZhangIESW18}, correlating well with human judgment. (4) \textbf{FID (Fréchet Inception Distance):} Measures distributional distance between reconstructed images in the feature space of an Inception network. Lower FID indicates higher quality. Appendix~\ref{apd:bootstrap_uncertainty} reports FID point estimates and bootstrap standard errors for four configurations on three 100-image datasets using 2,000 image-level resamples.

\textbf{Text Quality.}
Preserving text in images is critical for multimodal reasoning and OCR-related tasks. To quantify this, we apply OCR to the reconstructed images and compare the results to the OCR result of the original image using: (1) \textbf{CER (Character Error Rate):} The Levenshtein distance between predicted and ground truth characters, normalized by total character count. (2) \textbf{WER (Word Error Rate):} Similar to CER but computed over word sequences. It is particularly sensitive to segmentation and spelling accuracy. For OCR-based evaluation, we use Gemma 3~\citep{gemma3}, a SOTA multimodal model, to extract content from original and reconstructed images to calculate WER and CER. Details of OCR and WER/CER computation are included in Appendix~\ref{apd:text_quality_evaluation}.

These metrics allow us to analyze not only how much information is retained by the VT, but also what types of information (structural, perceptual, or symbolic) are lost in the tokenization process.

\begin{table*}
\centering
\caption{Evaluation of task 3 across three different settings.}\label{tbl:task3}
\resizebox{0.9\textwidth}{!}{%
\begin{tabular}{lc|cccccc|cccccc|cccccccc}
\toprule
\midrule
 &  & \multicolumn{6}{c|}{Movie Posters (Easy)}                                                                      & \multicolumn{6}{c|}{ArXiv Abstracts (Hard)}                                                                   & \multicolumn{8}{c}{Multilingual (Non-Latin)}                                                                                                                                                                                                                                                                                                                    \\
      {Method} & {Image Size}   & PSNR$\uparrow$   & SSIM$\uparrow$  & LPIPS$\downarrow$ & FID$\downarrow$  & CER$\downarrow$ & WER$\downarrow$ & PSNR$\uparrow$   & SSIM$\uparrow$  & LPIPS$\downarrow$ & FID$\downarrow$ & CER$\downarrow$ & WER$\downarrow$ & PSNR$\uparrow$   & SSIM$\uparrow$  & LPIPS$\downarrow$ & FID$\downarrow$  & \begin{tabular}[c]{@{}c@{}}CER$\downarrow$\\ (Chinese)\end{tabular} & \begin{tabular}[c]{@{}c@{}}CER$\downarrow$\\ (Hindi)\end{tabular} & \begin{tabular}[c]{@{}c@{}}CER$\downarrow$\\ (Japanese)\end{tabular} & \begin{tabular}[c]{@{}c@{}}CER$\downarrow$\\ (Korean)\end{tabular} \\\midrule
FlowMo Lo               & $256 \times 256$            & 24.1178          & 0.7471          & 0.0645            & 55.0685          & 0.6458          & 3.7022          & 14.4979          & 0.5961          & 0.0703            & 9.1933          & 0.8337          & 6.0968          & 13.3470          & 0.5694          & 0.1187            & 23.0079          & 1.2564                                                              & 1.0397                                                            & 1.3614                                                               & 1.2823                                                             \\
FlowMo Hi               & $256 \times 256$            & 27.4186          & 0.8403          & 0.0380            & 40.7009          & 0.4169          & 2.5639          & 15.8510          & 0.7262          & 0.0493            & 6.6921          & 0.7046          & 5.3013          & 14.4191          & 0.6932          & 0.0896            & 11.8131          & 1.1279                                                              & 0.9374                                                            & 1.0281                                                               & 0.9981                                                             \\
MaskBiT 16bit           & $256 \times 256$            & 21.0962          & 0.6748          & 0.0992            & 67.3479          & 0.7315          & 4.2234          & 14.3909          & 0.4616          & 0.1945            & 27.9940         & 1.3801          & 9.2146          & 13.2634          & 0.4157          & 0.3184            & 42.6548          & 3.1298                                                              & 1.1953                                                            & 2.0821                                                               & 2.1311                                                             \\
MaskBiT 18bit           & $256 \times 256$            & 21.2932          & 0.6876          & 0.0928            & 64.0766          & 0.7910          & 4.8894          & 14.7015          & 0.4978          & 0.2436            & 26.2086         & 1.2828          & 9.2956          & 13.4963          & 0.4450          & 0.3450            & 59.1847          & 3.4889                                                              & 1.1867                                                            & 1.9445                                                               & 2.3713                                                             \\
Titok-l32               & $256 \times 256$            & 15.5164          & 0.4595          & 0.2616            & 124.7412         & 1.2254          & 7.8547          & 12.2423          & 0.2390          & 0.2513            & 60.0889         & 1.1178          & 8.5836          & 11.6064          & 0.2440          & 0.3557            & 78.3890          & 3.0154                                                              & 1.1907                                                            & 1.9135                                                               & 1.7847                                                             \\
Titok-b64               & $256 \times 256$            & 16.9423          & 0.5273          & 0.1939            & 101.7892         & 1.0189          & 6.7070          & 12.9500          & 0.3068          & 0.2470            & 49.4931         & 1.2414          & 10.0394         & 11.7051          & 0.2557          & 0.3163            & 82.9833          & 3.5650                                                              & 1.4384                                                            & 2.0044                                                               & 2.3986                                                             \\
Titok-s128              & $256 \times 256$            & 17.0506          & 0.5427          & 0.1598            & 88.7466          & 1.2478          & 6.2260          & 12.9276          & 0.3297          & 0.2691            & 56.6373         & 1.4054          & 8.6644          & 11.7860          & 0.2710          & 0.3131            & 64.4152          & 2.7573                                                              & 1.1329                                                            & 1.6614                                                               & 1.9083                                                             \\
Titok-bl64              & $256 \times 256$            & 18.2597          & 0.5963          & 0.1584            & 95.2880          & 1.1421          & 6.7044          & 13.4414          & 0.4535          & 0.1398            & 26.4375         & 1.2385          & 9.9727          & 12.1882          & 0.3908          & 0.1801            & 38.8779          & 3.6473                                                              & 1.2105                                                            & 2.8315                                                               & 2.5622                                                             \\
Titok-bl128             & $256 \times 256$            & 19.8102          & 0.6502          & 0.1159            & 77.7980          & 1.5956          & 5.0851          & 13.5015          & 0.4810          & 0.1235            & 24.1506         & 1.0432          & 7.6730          & 12.5783          & 0.4306          & 0.2012            & 36.9834          & 2.3695                                                              & 0.9719                                                            & 2.1597                                                               & 1.8533                                                             \\
Titok-sl256             & $256 \times 256$            & 21.4568          & 0.7151          & 0.0841            & 61.0830          & 0.7850          & 4.5561          & 13.4954          & 0.4996          & 0.1122            & 19.0590         & 1.2131          & 8.2139          & 12.7078          & 0.4669          & 0.1589            & 26.0457          & 2.2980                                                              & 1.2338                                                            & 1.9775                                                               & 1.7088                                                             \\
OpenMagViT              & $256 \times 256$            & 22.5337          & 0.7440          & 0.0741            & 52.7153          & 0.7385          & 4.2419          & 13.6459          & 0.5472          & 0.0900            & 19.0211         & 1.0400          & 7.7046          & 12.3496          & 0.4874          & 0.1498            & 25.5786          & 2.5913                                                              & 1.1622                                                            & 2.0300                                                               & 1.9826                                                             \\
LlamaGen ds8            & $256 \times 256$            & 26.0990          & 0.8288          & 0.0371            & 31.4534          & 0.5066          & 2.9620          & 16.0494          & 0.7003          & 0.0648            & 25.1991         & 0.9769          & 7.1170          & 14.5250          & 0.6513          & 0.1024            & 19.2362          & 1.1408                                                              & 1.0250                                                            & 1.0505                                                               & 1.1342                                                             \\
BSQ-VIT                 & $256 \times 256$            & \textbf{30.7305} & \textbf{0.9135} & \textbf{0.0172}   & \textbf{25.6235} & \textbf{0.1102} & \textbf{0.5763} & \textbf{20.3628} & \textbf{0.8910} & \textbf{0.0270}   & \textbf{4.7719} & \textbf{0.0459} & \textbf{0.3234} & \textbf{17.9849} & \textbf{0.8544} & \textbf{0.0582}   & \textbf{10.2721} & \textbf{0.7858}                                                     & \textbf{0.4722}                                                   & \textbf{0.3827}                                                      & \textbf{0.5919}                                                    \\
VAR-256                 & $256 \times 256$            & 23.6215          & 0.7349          & 0.0602            & 50.2971          & 0.6391          & 3.7266          & 15.2481          & 0.6799          & 0.0569            & 23.1849         & 1.0170          & 7.3612          & 13.1452          & 0.5672          & 0.1165            & 19.0211          & 2.3378                                                              & 1.1118                                                            & 1.8972                                                               & 1.5754                                                             \\\midrule
Janus Pro 1B/7B         & $384 \times 384$            & 25.2793          & 0.8203          & 0.0545            & 45.4080          & 0.6628          & 3.7549          & 17.0286          & 0.6535          & 0.1003            & 38.3604         & 1.0645          & 7.8280          & 15.7220          & 0.6223          & 0.1367            & 53.0056          & 1.5413                                                              & 1.1578                                                            & 1.3463                                                               & 1.3333                                                             \\\midrule
Chameleon               & $512 \times 512$            & 27.1939          & 0.8358          & 0.0430            & 39.8786          & 0.5782          & 3.1359          & 18.1545          & 0.7503          & 0.0588            & 8.1713          & 0.7158          & 5.1867          & 15.3719          & 0.6370          & 0.0977            & \textbf{15.5254} & 1.9706                                                              & 0.9171                                                            & 1.5444                                                               & 1.4047                                                             \\
LlamaGen ds16           & $512 \times 512$            & 27.3778          & \textbf{0.8717} & 0.0440            & 40.9660          & 0.4914          & 2.8774          & 19.0951          & 0.7601          & 0.0601            & 18.1659         & 0.9527          & 6.8741          & 17.2316          & 0.7068          & 0.1141            & 38.2967          & 1.2307                                                              & 1.0938                                                            & 1.0343                                                               & 1.1108                                                             \\
LlamaGen ds16 T2I       & $512 \times 512$            & \textbf{27.4929} & 0.8697          & \textbf{0.0411}   & \textbf{38.2986} & 0.4842          & 2.5741          & 18.8911          & 0.7735          & 0.0567            & 20.8214         & 0.9010          & 6.1850          & 17.2267          & 0.7400          & \textbf{0.0869}   & 45.0646          & 0.9993                                                              & 1.0305                                                            & 0.8139                                                               & 0.9775                                                             \\
VAR-512                 & $512 \times 512$            & 26.4574          & 0.8331          & 0.0486            & 46.6674          & \textbf{0.1949} & \textbf{1.1867} & \textbf{21.0023} & \textbf{0.8561} & \textbf{0.0553}   & \textbf{5.3145} & \textbf{0.1341} & \textbf{0.9829} & \textbf{17.7790} & \textbf{0.7687} & 0.1012            & 22.1121          & \textbf{0.8398}                                                     & \textbf{0.6873}                                                   & \textbf{0.6552}                                                      & \textbf{0.7067}                                                    \\\midrule
Infinity-d32            & $1024 \times 1024$          & 37.7533          & 0.9669          & 0.0063            & 10.5845          & 0.0790          & 0.4371          & 31.7028          & 0.9896          & 0.0049            & 0.7982          & 0.0017          & 0.0162          & 30.2736          & 0.9892          & 0.0086            & 2.2108           & 0.1758                                                              & 0.0352                                                            & 0.0661                                                               & 0.0924                                                             \\
Infinity-d64            & $1024 \times 1024$          & 39.3334          & 0.9774          & 0.0048            & 8.6894           & 0.0439          & 0.2978          & 35.0924          & 0.9952          & 0.0026            & 0.7426          & 0.0019          & 0.0165          & 34.5168          & 0.9959          & 0.0035            & 1.1291           & 0.1372                                                              & 0.0604                                                            & 0.0617                                                               & 0.0582                                                             \\\midrule
SD3.5L                  & $1024 \times 1024$          & 38.9058          & 0.9650          & 0.0011            & 1.6215           & \textbf{0.0002} & \textbf{0.0000} & 40.1536          & 0.9988          & 0.0005            & 0.1331          & \textbf{0.0001} & \textbf{0.0010} & 39.6703          & 0.9989          & 0.0011            & 0.4205           & \textbf{0.0973}                                                     & 0.0310                                                            & 0.1047                                                               & 0.0545                                                             \\
FLUX.1-dev              & $1024 \times 1024$          & \textbf{44.3863} & \textbf{0.9862} & \textbf{0.0005}   & \textbf{0.8993}  & 0.0521          & 0.4872          & \textbf{52.1000} & \textbf{0.9999} & \textbf{0.0001}   & \textbf{0.0256} & 0.0018          & 0.0112          & \textbf{51.9243} & \textbf{0.9999} & \textbf{0.0001}   & \textbf{0.0457}  & 0.1220                                                              & \textbf{0.0272}                                                   & \textbf{0.0614}                                                      & \textbf{0.0506}                                                    \\
GPT-4o (ref.)                  & $1024 \times 1024$          & 17.7677          & 0.6006          & 0.1388            & 63.7351          & 0.0180          & 0.0836          & 11.3299          & 0.1516          & 0.1929            & 32.5382         & 0.4928          & 3.5292          & 10.4376          & 0.1761          & 0.2291            & 20.0185          & 0.8096                                                              & 0.6414                                                            & 0.7219                                                               & 0.6478                                                            
            \\
\midrule
\bottomrule
\end{tabular}%
}
\end{table*}

\section{Experimental Results \& Analysis}

\textbf{Models.}
We evaluate a diverse set of VTs spanning multiple quantization paradigms, including VQ, LFQ, BSQ, RVQ, and RBSQ. Our benchmark covers a wide range of SOTA models such as FlowMo~\citep{flowmo}, MaskBiT~\citep{maskbit}, Titok~\citep{titok}, OpenMagViT2~\citep{openmagvit2}, LlamaGen~\citep{llamagen}, BSQ-ViT~\citep{BSQViT}, VAR~\citep{VAR}, Janus Pro~\citep{januspro}, Chameleon~\citep{Chameleon}, and Infinity~\citep{infinity}. In our experiments, we specifically evaluate the VT component in isolation, without modifying or including the downstream generation models. This design choice allows us to focus purely on the tokenizer’s ability to preserve visual information. The released tokenizer configurations differ in compression ratio, latent dimensionality, training data and objectives, and encoder--decoder design. For comparison, we also include continuous VAEs used in diffusion models (e.g., SD3.5L~\citep{stable-diffusion}, FLUX.1-dev~\citep{flux2024}) and provide results from GPT-4o~\citep{openai2025image} as a system-level reference (ref.), although its tokenizer architecture remains undisclosed~\citep{yan2025gpt}. GPT-4o-generated images are fixed evaluation inputs shared across methods. For GPT-4o, we use the prompt ``\textit{Please recreate the exact same image without any alterations and preserve the original resolution.}'' to reconstruct images. Table~\ref{tbl:task1} summarizes all evaluated models with their parameter counts and quantization types. Detailed experimental settings and environments are reported in Appendix~\ref{apd:experimental_setting}. Additional SBER-MoVQGAN\footnote{Official repository: \url{https://github.com/ai-forever/MoVQGAN}.} results for 67M/102M/270M on the 100-image detail, poster, and English-abstract subsets appear in Appendix~\ref{apd:sber_movqgan}.

\begin{table}[t]
\centering
\caption{Evaluation of VTs on task 2.}\label{tbl:task2}
\resizebox{.43\textwidth}{!}{%
\begin{tabular}{lccccc}
\toprule
\midrule
\multicolumn{1}{l}{Method} & Image Size         & \multicolumn{1}{c}{PSNR$\uparrow$} & \multicolumn{1}{c}{SSIM$\uparrow$} & \multicolumn{1}{c}{LPIPS$\downarrow$} & \multicolumn{1}{c}{FID$\downarrow$} \\
\midrule
FlowMo Lo                  & $256 \times 256$   & 17.0642                            & 0.3578                             & 0.1344                                & 64.2895                             \\
FlowMo Hi                  & $256 \times 256$   & 19.0346                            & 0.5578                             & 0.0845                                & 44.5536                             \\
MaskBiT 16bit              & $256 \times 256$   & 17.2804                            & 0.2644                             & 0.1803                                & 98.5133                             \\
MaskBiT 18bit              & $256 \times 256$   & 17.3603                            & 0.2728                             & 0.1745                                & 91.3108                             \\
Titok-l32                  & $256 \times 256$   & 15.0325                            & 0.1609                             & 0.3474                                & 169.2162                            \\
Titok-b64                  & $256 \times 256$   & 15.3827                            & 0.1666                             & 0.2767                                & 124.9069                            \\
Titok-s128                 & $256 \times 256$   & 15.6532                            & 0.1828                             & 0.2316                                & 107.7550                            \\
Titok-bl64                 & $256 \times 256$   & 15.7606                            & 0.1894                             & 0.2334                                & 121.5447                            \\
Titok-bl128                & $256 \times 256$   & 16.3017                            & 0.2373                             & 0.1808                                & 93.1151                             \\
Titok-sl256                & $256 \times 256$   & 17.1632                            & 0.3091                             & 0.1413                                & 69.7221                             \\
OpenMagViT                 & $256 \times 256$   & 17.0070                            & 0.3278                             & 0.1406                                & 70.6742                             \\
LlamaGen ds8               & $256 \times 256$   & 19.1859                            & 0.5144                             & 0.0852                                & 42.8318                             \\
BSQ-VIT                    & $256 \times 256$   & \textbf{22.4836}                   & \textbf{0.7487}                    & \textbf{0.0496}                       & \textbf{28.5461}                    \\
VAR-256                    & $256 \times 256$   & 17.8223                            & 0.3960                             & 0.1207                                & 53.6363                             \\
\midrule
Janus Pro 1B/7B            & $384 \times 384$   & 20.4726                            & 0.4908                             & 0.1174                                & 59.6669                             \\
\midrule
Chameleon                  & $512 \times 512$   & 20.5686                            & 0.5089                             & 0.1182                                & 67.0157                             \\
LlamaGen ds16              & $512 \times 512$   & 21.8902                            & 0.5903                             & 0.0972                                & 51.3484                             \\
LlamaGen ds16 T2I          & $512 \times 512$   & \textbf{21.9694}                   & 0.6029                             & \textbf{0.0947}                       & \textbf{51.0251}                    \\
VAR-512                    & $512 \times 512$   & 21.5492                            & \textbf{0.6537}                    & 0.1000                                & 57.3488                             \\
\midrule
Infinity-d32               & $1024 \times 1024$ & 29.1329                            & 0.9052                             & 0.0269                                & 21.0871                             \\
Infinity-d64               & $1024 \times 1024$ & 31.1374                            & 0.9396                             & 0.0188                                & 16.7424                             \\\midrule
SD3.5L                     & $1024 \times 1024$ & 37.4365                            & 0.9774                             & 0.0028                                & 4.5717                              \\
FLUX.1-dev                 & $1024 \times 1024$ & \textbf{41.5280}                   & \textbf{0.9908}                    & \textbf{0.0011}                       & \textbf{2.1202}                     \\
GPT-4o (ref.)                     & $1024 \times 1024$ & 15.8109                            & 0.3000                             & 0.2156                                & 75.8916                          \\

\midrule
\bottomrule
\end{tabular}%
}
\end{table}

\subsection{Task 1: Image Reconstruction}

In this task, we evaluate image reconstruction across three subtasks: (1) ImageNet: Images are center-cropped and resized to match each model's required input size. (2) High Resolution: All images are set to a fixed size of $1024 \times 1024$, and we retain this resolution for all models without resizing. 
(3) Varying Resolution:  This subset contains images with diverse, non-uniform resolutions, and we preserve their original sizes for evaluation. We assess reconstruction quality using PSNR, SSIM, LPIPS, and FID. 
The results in Table~\ref{tbl:task1} show that continuous VAEs achieve stronger reconstruction fidelity than discrete VTs, including the same-resolution comparisons at $1024 \times 1024$. Notably, results under our direct whole-image High Resolution and Varying Resolution protocol highlight that most VTs are limited to model-specific input sizes and fail to generalize to arbitrary resolutions, unlike continuous VAEs, which naturally support flexible image dimensions.

\paragraph{Reconstruction and generation.} Within-family comparisons provide evidence for the link between reconstruction and generation quality. TiTok L-32, B-64, and S-128 have VTBench reconstruction FIDs of 1.818, 1.354, and 1.419, alongside published generation FIDs of 2.77, 2.48, and 2.50~\citep{titok}; the two rankings agree. For MaskBiT, 18-bit has a lower VTBench reconstruction FID than 16-bit (1.033 versus 1.147) and a slightly higher released generation FID (1.67 versus 1.64)~\citep{maskbit}.\footnote{Generation FIDs are from the official MaskBiT model zoo: \url{https://github.com/markweberdev/maskbit}.} These patterns motivate evaluating reconstruction fidelity alongside tokenizer--generator compatibility. Generation FIDs use the authors' published configurations: TiTok uses 8 sampling steps with MaskGIT-ViT (L-32/B-64) or MaskGIT-UViT-L (S-128), and MaskBiT uses 64 steps.

Figure~\ref{fig:task1_examples} shows qualitative reconstructions across VTs. We observe consistent semantic degradations in discrete VT outputs. For example, red crosses on ambulances are missing or blurred (ImageNet), rooftops are distorted, and facial details and expressions are noticeably altered (High Resolution). In the Varying Resolution, background textures and object boundaries become heavily corrupted. Under this whole-image protocol, models such as Janus Pro, LlamaGen, Chameleon, and FlowMo do not raise runtime errors but fail to generate correct semantic content, highlighting the inherent limitations of current VTs, especially under high or non-regular resolutions.

\subsection{Task 2: Detail Preservation}
In this task, we evaluate the ability of each VT to retain high-frequency visual information using a dataset of patterned and texture-rich images. As shown in Table~\ref{tbl:task2}, continuous VAEs again lead in all metrics. Among discrete VTs, Infinity-d64 achieves the strongest detail-preservation results. We further visualize qualitative results in Figure~\ref{fig:task2_examples} (Appendix~\ref{apd:additional_visualization}). While continuous models such as SD3.5L, and FLUX.1 closely resemble the original, many VTs introduce visible distortions, such as blurred lines, broken shapes, and color bleeding. These results highlight the limitations of current discrete tokenization schemes in retaining local textures and structural integrity.

\subsection{Task 3: Text Preservation}
This task evaluates how well VTs preserve text content in images -- a crucial capability for OCR, document understanding, and multimodal reasoning. We consider three increasingly challenging scenarios: (1) Movie Posters with clean, short English text; (2) ArXiv Abstracts containing dense, long-form academic writing; and (3) Multilingual Text rendered in non-Latin scripts (Chinese, Hindi, Japanese, Korean). We use Character Error Rate (CER) and Word Error Rate (WER), based on OCR outputs from the Gemma 3 model, to quantify text preservation. As shown in Table~\ref{tbl:task3} and Figure~\ref{fig:task2_examples} (Appendix~\ref{apd:additional_visualization}), continuous VAEs achieve strong text-preservation results across the evaluated settings. Notably, many VTs fail to reconstruct slogans or titles in the Movie Poster setting, and perform worse on academic layouts or multilingual scripts. These results highlight the limitations of discrete tokenization methods in preserving symbolic fidelity, particularly under complex formatting and linguistic diversity. Additional experiments with alternative OCR backends are reported in Appendix~\ref{apd:ocr_sensitivity}, supporting these findings across OCR systems.

\paragraph{SBER-MoVQGAN.} PSNR, SSIM, and LPIPS improve across all three subtasks as model size increases; text-error rates also decrease. English abstract FID favors 102M (3.9158 vs. 4.2655 for 270M; Appendix~\ref{apd:sber_movqgan}).

\subsection{Failure Modes and Design Implications}
\paragraph{Capacity and symbolic fidelity.} TiTok-BL128 improves ImageNet LPIPS over BL64 (0.147 vs. 0.193); Infinity-d64 improves detail LPIPS over d32 (0.0188 vs. 0.0269). FlowMo Hi and VAR-512 improve all eight text-error rates over Lo and VAR-256. MaskBiT 18-bit improves reconstruction and detail metrics but worsens poster CER/WER (0.7910/4.8894 vs. 0.7315/4.2234 for 16-bit). Symbolic fidelity depends on more than nominal~capacity.
\paragraph{Whole-image resolution failures.} FlowMo Hi's whole-image failures are associated with lower capacity per pixel and evaluation outside its training resolution (Appendix~\ref{apd:experimental_setting}).
\paragraph{Design implications.} These results motivate greater effective capacity, flexible latent layouts and decoders for high and varying resolutions, and detail- and text-aware objectives.

\section{Conclusion}
In this paper, we introduce \textbf{VTBench}, a benchmark for systematically evaluating visual tokenizers in autoregressive image generation. VTBench covers three tasks: image reconstruction, detail preservation, and text preservation. Experiments across major visual-tokenization paradigms show a clear overall advantage of continuous VAEs over discrete VTs in reconstruction fidelity, spatial consistency, and text accuracy. VTBench facilitates the development of stronger and more general-purpose open-source visual tokenizers, advancing unified multimodal generative modeling.

\clearpage

\section*{Limitations}
VTBench focuses on evaluating visual tokenizers in isolation rather than as part of an end-to-end generation pipeline; downstream interactions between a tokenizer and an autoregressive backbone are therefore not directly measured. Our text-preservation evaluation relies on OCR systems, which are themselves imperfect; although Appendix~\ref{apd:ocr_sensitivity} supports consistent overall performance trends across OCR backends, absolute CER/WER values may still depend on the chosen OCR model. The benchmark covers a representative but not exhaustive set of VTs, and proprietary tokenizers such as GPT-4o's are evaluated only through their public APIs. Finally, the multilingual subtask currently includes four non-Latin scripts (Chinese, Hindi, Japanese, Korean); broader linguistic coverage and additional domain-specific scenarios are left to future work. The targeted 100-image subsets provide focused coverage of specific visual challenges. Domain-specific imagery, noise and photometric corruptions, and geometric perturbations remain future extensions. Prompt-injection success~\citep{DBLP:journals/corr/abs-2502-13141} requires evaluation with a downstream multimodal model in addition to measuring embedded-text preservation. LLMs were used solely to polish and revise the writing; the human authors are fully responsible for all claims, results, and conclusions in this paper.

\section*{Acknowledgements}
This work is supported in part by the National Science Foundation (NSF) under Award Nos. 2247619 and 2543795. This work is also supported by NSF under Award No. 2610649 and No. 2326494 and by NERSC through DDR-ERCAP0035256.
We thank Research Computing at \citet{rit_research_computing_2026} for providing computational resources and support for this work.

\bibliography{references}

\appendix

\section{Extended Background on Visual Tokenizers}\label{apd:vt-background}
While the main text provides a brief overview of VTs, this section offers a more detailed explanation of representative visual tokenizers, including both continuous and discrete variants. We present formal definitions and architectural insights for each method, laying the groundwork for understanding their differences in performance and suitability for autoregressive image generation.

\textbf{Continuous VAE.} In diffusion models, a common approach is to use a continuous Variational Autoencoder (VAE)~\citep{stable-diffusion, DBLP:conf/iccv/PeeblesX23} as a feature compressor as shown in Figure~\ref{fig:VT_framework}(a). These VAEs encode the input image into a continuous latent space, typically using convolutional encoders and decoders, enabling high-quality image reconstructions. Formally, let $x \in \mathbb{R}^{H \times W \times 3}$ be the input image, where $H$ and $W$ denote the image’s height and width, respectively. The encoder $E$ maps $x$ to a latent representation $z = E(x) \in \mathbb{R}^{h \times w \times d}$, where $h$ and $w$ are the spatial dimensions of the latent space (typically downsampled from $H$ and $W$), and $d$ is the latent channel dimension. The decoder $D$ reconstructs the image as $\hat{x} = D(z)$. However, AR models require inputs in the form of discrete token sequences, motivating the development of discrete visual tokenizers.

\textbf{VQ-VAE.} Vector Quantized Variational Autoencoder (VQ-VAE)~\citep{vqvae} is one of the earliest and most widely adopted discrete visual tokenizers. It encodes an image into a latent feature map and then quantizes each spatial location to the closest vector in a learned codebook. The resulting discrete indices form a token grid that can be used in AR models. Given an input image $x \in \mathbb{R}^{H \times W \times 3}$, the encoder $E$ produces a latent map $z = E(x) \in \mathbb{R}^{h \times w \times d}$. Each latent vector $z_{i,j}$ is replaced by its nearest neighbor in the codebook $\mathcal{C} = \{c_1, c_2, \dots, c_K\} \subset \mathbb{R}^d$, $q_{i, j} = \arg\min_{k}||z_{i,j}-c_k||^2_2,\ \hat{z}_{i,j}=c_{q_{i,j}}$. The quantized latent $\hat{z}$ is then decoded via $D$ to reconstruct the image $\hat{x} = D(\hat{z})$. While VQ-VAE enables discrete tokenization suitable for AR models, it suffers from limited expressivity, codebook collapse, and scalability issues when handling high-resolution or diverse image content.

\textbf{LFQ-VAE.} Lookup-Free Quantization (LFQ)~\citep{LFQ} eliminates the nearest-neighbor lookup and codebook structure by directly projecting latent features into a binary latent space using a learnable mapping. The encoder $E$ produces $z = E(x) \in \mathbb{R}^{h \times w \times d}$. Each element $z_{i,j,k}$ is quantized using a sign function: $\hat{z}_{i,j,k} = \text{sign}(z_{i,j,k})\in\{-1,+1\}$. The binary representation can be interpreted as a discrete token mask $m$, where each binary vector at location $(i,j)$ is mapped to an integer token using $m_{i,j} = \sum_{k=1}^{d} 2^{k-1} \one_{z_{i,j,k} > 0}$.

\textbf{BSQ-VAE.} Binary Spherical Quantization (BSQ)~\citep{BSQViT} builds upon LFQ by incorporating $\ell_2$ normalization before quantization: $\hat{z}_{i,j,k} = \frac{1}{|\sqrt{d}|}\text{sign}(\frac{z_{i,j,k}}{|z_{i,j}|})$.This constrains the representation to a unit hypersphere, effectively reducing quantization error and leading to smoother latent spaces. BSQ allows for finer-grained tokenization while maintaining efficient binary encoding, which is beneficial for downstream tasks sensitive to visual detail.

\textbf{Residual Next-Scale VAE} introduces hierarchical tokenization by progressively quantizing residual information across multiple spatial scales~\citep{VAR}. Let $s \in \{1, 2, \dots, n\}$ denote the scale index, where $n$ is the total number of scales. Each scale $s$ corresponds to a predefined spatial resolution $(h_s, w_s)$, with $h_s \leq h$ and $w_s \leq w$. During encoding, a sequence of quantized tokens $\hat{z} = \{\hat{z}^{(1)}, \hat{z}^{(2)}, \dots, \hat{z}^{(n)}\}$ is computed as follows:

\begin{equation}
\left\{
\begin{aligned}
& z = r^{(0)} = E(x) \in \mathbb{R}^{h \times w \times d} \\
& \hat{z}^{(s)} = \mathcal{Q}\!\left(\downarrow_{(h_s, w_s)}\!\left(r^{(s-1)}\right)\right), \\
& \qquad r^{(s)} = z - \uparrow_{(h, w)}\!\left(\mathcal{D}\!\left(\hat{z}^{(s)}\right)\right), \quad s = 1,\dots,n
\end{aligned}
\right.
\end{equation}

Here, $\mathcal{Q}$ and $\mathcal{D}$ represent the quantization and dequantization functions, respectively, which may be implemented using VQ, LFQ, BSQ, or similar methods. The operators $\downarrow_{(h_s, w_s)}(\cdot)$ and $\uparrow_{(h, w)}(\cdot)$ denote downsampling to and upsampling from the specified resolution. During decoding, the final image is reconstructed by aggregating and decoding all quantized components:
\begin{align}
\hat{x} = D\left( \sum_{s=1}^{n} \uparrow_{(h, w)}\left( \mathcal{D}(\hat{z}^{(s)}) \right) \right)
\end{align}
This coarse-to-fine scheme preserves spatial structure and enables the model to capture both global context and fine-grained details, making it effective for high-resolution image reconstruction.

\section{Definition of Visual Details}
\label{apd:detail_definition}

The goal of Task 2 is to evaluate the ability of VTs to retain image details. In this section, we define what is meant by ``details'' in the context of Task 2.

\textit{\textbf{What Is ``Detail''?}} In the context of VTBench, we define ``visual detail'' as high-frequency visual information that is essential for preserving the local semantic integrity of an image. Concretely, we categorize details into the following visual elements: (1) Fine textures (e.g., fabric weave, hair strands, foliage patterns); and (2)Sharp edges and contours (e.g., architectural lines, object boundaries). These are typically spatially localized, semantically significant, and vulnerable to information loss during quantization. Unlike global semantics (e.g., category labels), these details are often non-redundant and not easily recoverable once degraded, especially in AR generation pipelines where token quality directly determines synthesis fidelity.

\textit{\textbf{Why This Definition Matters for Tokenization?}} Discrete VT compress image features into finite token vocabularies. During this compression, high-frequency information is the first to be lost, particularly under quantization or low expressiveness of token representations. This directly impacts downstream generation quality by perceptual degradation (textures appear smoothed, edges blurred, and small features missed) Therefore, evaluating a VT's ability to retain such localized, semantically relevant details is essential for both perceptual quality and task-level accuracy in multimodal applications.

Unlike global semantic content, such as scene category or layout, these local features are often non-redundant and highly sensitive to quantization errors. Preserving them is essential for perceptual quality and for downstream tasks that rely on accurate image content.

To make this definition operational and measurable, we designed Task 2: Detail Preservation specifically to target and evaluate a model's ability to retain such features. To validate that Task 2 indeed focuses on detail-rich content, we computed three standard image analysis metrics across all tasks as show in Table~\ref{tbl:detail_evaluation_all_tasks}:

\begin{itemize}[itemsep=2pt, topsep=0pt, leftmargin=20pt, parsep=0pt, partopsep=0pt]
\item \textbf{High-Frequency Energy (\%):} This measures the proportion of image energy in high-frequency components (computed via FFT), indicating the global presence of textures and patterns.
\item \textbf{Laplacian Variance:} This quantifies the overall sharpness of an image by measuring the variance of the Laplacian response, a standard proxy for edge clarity and detail strength.
\item \textbf{Edge Density (\%):} This captures the spatial density of edges in an image, revealing how concentrated and frequent fine structures are within a scene.
\end{itemize}

\begin{table*}[h]
\centering
\caption{Verification that Task 2 images contain richer details.}\label{tbl:detail_evaluation_all_tasks}
\resizebox{.85\textwidth}{!}{%
\begin{tabular}{llccc}
\toprule
\midrule
Task                    & Subtask                      & \multicolumn{1}{l}{HighFreq Energy(\%) $\uparrow$} & \multicolumn{1}{l}{Laplacian Variance $\uparrow$} & \multicolumn{1}{l}{Edge Density(\%) $\uparrow$} \\
\midrule
\multirow{3}{*}{Task 1} & Imagenet                     & 40.7911                                            & 2735.9673                                         & 11.0883                                         \\
                        & High Resolution              & 40.3702                                            & 940.7714                                          & 5.0195                                          \\
                        & Varying Resolution           & 34.1957                                            & 1188.534                                          & 9.3666                                          \\
\midrule
Task 2                  & \textbf{Detail Preservation} & \textbf{48.931}                                    & \textbf{12852.3598}                               & \textbf{23.6103}                                \\
\midrule
\multirow{6}{*}{Task 3} & Movie Posters                & 44.2046                                            & 1288.8568                                         & 2.3493                                          \\
                        & Arxiv Abstracts              & 39.6323                                            & 9857.1757                                         & 9.1615                                          \\
                        & Multilingual (Chinese)       & 31.5059                                            & 8212.4345                                         & 10.1427                                         \\
                        & Multilingual (Hindi)         & 31.7221                                            & 8087.6291                                         & 9.0816                                          \\
                        & Multilingual (Japanese)      & 32.3393                                            & 8691.5516                                         & 9.6565                                          \\
                        & Multilingual (Korean)        & 29.983                                             & 6847.8147                                         & 9.5393                                         
 \\
\midrule
\bottomrule
\end{tabular}%
}
\end{table*}

As shown in Table~\ref{tbl:detail_evaluation_all_tasks}, among all benchmark tasks, Task 2 exhibits the highest values across all three indicators. Specifically, it has a High-Frequency Energy of 48.93\%, a Laplacian Variance of 12,852.36, and an Edge Density of 23.61\%. These values far exceed those of Task 1 (ImageNet-Val), which show only 40.79\%, 2735.97, and 11.09\%, respectively, as well as all variants of Task 3. For example, even the ArXiv Abstracts test set, which contains visually complex scientific documents, reaches only 39.63\% High-Frequency Energy and 9.16\% Edge Density.

This quantitative evidence demonstrates that Task 2 is constructed to challenge models on their ability to preserve fine-grained, localized, and perceptually significant visual features. By combining these measurements with perceptual similarity metrics such as LPIPS in evaluation, we ensure a rigorous and reproducible assessment of detail preservation quality.

\section{Resolution Limitations of Discrete VTs}
\label{apd:resolution_limitations}

Although discrete visual tokenizers (VTs) and continuous variational autoencoders (VAEs) share a similar high-level encoder–decoder paradigm, their behavior diverges substantially when handling arbitrary image resolutions. In practice, current discrete VTs exhibit several architectural and training limitations that hinder their ability to generalize across resolutions, in contrast to the fully convolutional structure of continuous VAEs.

\paragraph{Input Size Constraints.} Many discrete VTs, particularly those based on vector quantization (e.g., VQ-VAE, LFQ), require fixed input sizes during both training and inference. This constraint arises from codebook usage, patchification schemes, and decoder assumptions (e.g., flattening 2D grids into token sequences of fixed length). For example, models such as Janus support only predefined square resolutions (e.g., $256 \times 256$, $384 \times 384$, or $512 \times 512$). Processing unseen resolutions often results in runtime errors or corrupted outputs, as illustrated in Figure~\ref{fig:task1_examples}.

\paragraph{Decoder Design Dependency.} The decoders of discrete VTs typically assume a token grid shape to map sequences back into spatial layouts. This coupling binds input resolution to the decoder architecture. In contrast, convolution-based VAEs, such as those employed in diffusion models, naturally support arbitrary resolutions because of their fully convolutional encoder–decoder design.

\paragraph{Token Flattening and Positional Encoding.} Autoregressive models built on discrete tokens usually flatten spatial grids into one-dimensional sequences. This flattening exacerbates resolution sensitivity: token grid dimensions change with input size, but positional embeddings and transformer blocks are rarely designed to handle such variation. Consequently, discrete VTs exhibit reduced robustness when applied to arbitrary resolutions.

In summary, while quantization itself is not inherently resolution-sensitive, current implementations of discrete VTs are limited in practice by input size assumptions, decoder rigidity, and positional encoding designs. These architectural constraints make discrete VTs significantly less robust to varying resolutions compared to continuous VAEs.

\section{Text Quality Evaluation}
\label{apd:text_quality_evaluation}

Evaluating the ability of a VT to preserve text content is critical for many downstream applications such as OCR, document generation/understanding, and multimodal reasoning. In VTBench, we quantify this ability by applying Optical Character Recognition (OCR) to both original and reconstructed images, and computing standard text similarity metrics based on the OCR outputs.

\textbf{Optical Character Recognition (OCR)} We use a SOTA multimodal model, Gemma 3~\citep{gemma3}, to extract textual content from both the original image $x$ and the reconstructed image $\hat{x}$. Let $T_{\text{orig}}$ and $T_{\text{recon}}$ denote the sequences obtained from $x$ and $\hat{x}$, respectively. All evaluations are performed under the same OCR configuration to ensure consistency.

\textbf{Character Error Rate (CER)} The Character Error Rate measures the normalized edit distance between two character sequences. Given the ground-truth sequence $T_{\text{orig}}$ and the predicted sequence $T_{\text{recon}}$, CER is computed as:

\begin{align}
\text{CER} = \frac{D_{\text{char}}(T_{\text{orig}}, T_{\text{recon}})}{|T_{\text{orig}}|}
\end{align}

where $D_{\text{char}}(\cdot, \cdot)$ denotes the Levenshtein distance (i.e., the minimum number of insertions, deletions, and substitutions needed to convert one character sequence to another), and $|T_{\text{orig}}|$ is the number of characters in the ground-truth sequence.

\textbf{Word Error Rate (WER).} Word Error Rate is similar to CER, but computed at the word level. Let $W_{\text{orig}}$ and $W_{\text{recon}}$ be the sequences of words extracted from the original and reconstructed images, respectively. The WER is defined as:

\begin{align}
\text{WER} = \frac{D_{\text{word}}(W_{\text{orig}}, W_{\text{recon}})}{|W_{\text{orig}}|}
\end{align}

where $D_{\text{word}}(\cdot, \cdot)$ is the Levenshtein distance between word sequences, and $|W_{\text{orig}}|$ is the number of words in the ground-truth text.

As with CER, WER can also exceed 1 in extreme failure cases, where the number of incorrect or missing words far surpasses the original word count. These metrics jointly capture both low-level character accuracy and higher-level semantic correctness. High CER or WER indicates that the visual tokenizer has failed to preserve text layout or visual clarity during reconstruction, especially in multilingual or high-density text scenarios.

\section{Robustness of Text Preservation to OCR Choice}
\label{apd:ocr_sensitivity}

Our text preservation evaluation in Task~3 relies on OCR outputs to compute character error rate (CER) and word error rate (WER). A natural concern is whether the results are sensitive to the choice of OCR system. To investigate this, we conducted a follow-up experiment with two additional OCR backends beyond Gemma~3: \texttt{Qwen2.5-VL-7B-Instruct} and \texttt{nanonets/Nanonets-OCR-s}. The CER/WER results across all three OCR systems are summarized in Table~\ref{tab:ocr_comparison}.

\begin{table}[h]
\centering
\caption{Comparison of OCR sensitivity in text preservation evaluation. Overall performance trends are consistent across Gemma~3, Qwen2.5-VL-7B-Instruct, and Nanonets-OCR-s.}
\resizebox{.5\textwidth}{!}{%
\begin{tabular}{l|cc|cc|cc}
\toprule
\midrule
\multirow{2}{*}{Method} & \multicolumn{2}{c|}{Gemma~3} & \multicolumn{2}{c|}{Qwen2.5-VL-7B-Instruct} & \multicolumn{2}{c}{Nanonets-OCR-s} \\
 & CER & WER & CER & WER & CER & WER \\
\midrule
FlowMo Lo & 0.6458 & 3.7022 & 0.5969 & 3.4291 & 0.5861 & 3.3744 \\
FlowMo Hi & 0.4169 & 2.5639 & 0.4148 & 2.3856 & 0.3987 & 2.2985 \\
MaskBiT 16bit & 0.7315 & 4.2234 & 0.7679 & 4.4453 & 1.0437 & 6.0299 \\
MaskBiT 18bit & 0.7910 & 4.8894 & 0.7599 & 4.3905 & 0.7330 & 4.2090 \\
Titok-l32 & 1.2254 & 7.8547 & 0.9549 & 5.5398 & 2.1851 & 10.1903 \\
Titok-b64 & 1.0189 & 6.7070 & 1.1570 & 5.5000 & 1.3111 & 7.5522 \\
Titok-s128 & 1.2478 & 6.2260 & 0.9538 & 8.1082 & 1.1665 & 6.7177 \\
Titok-bl64 & 1.1421 & 6.7044 & 1.1605 & 6.7313 & 1.9725 & 6.4577 \\
Titok-bl128 & 1.5956 & 5.0851 & 1.2116 & 7.8060 & 2.6343 & 15.3209 \\
Titok-sl256 & 0.7850 & 4.5561 & 0.7634 & 4.4179 & 0.7377 & 4.2376 \\
OpenMagViT & 0.7385 & 4.2419 & 0.6932 & 4.0100 & 0.8561 & 4.9428 \\
LlamaGen ds8 & 0.5066 & 2.9620 & 0.5471 & 3.1679 & 0.4913 & 2.8445 \\
BSQ-VIT & 0.1102 & 0.5763 & 0.1059 & 0.6480 & 0.0659 & 0.3495 \\
VAR-256 & 0.6391 & 3.7266 & 0.6243 & 3.6206 & 0.6130 & 3.5448 \\
Janus Pro 1B/7B & 0.6628 & 3.7549 & 0.6488 & 3.7920 & 1.0328 & 4.7400 \\
Chameleon & 0.5782 & 3.1359 & 0.5369 & 3.1295 & 0.7092 & 4.0871 \\
LlamaGen ds16 & 0.4914 & 2.8774 & 0.5112 & 2.9676 & 0.4704 & 2.6841 \\
LlamaGen ds16 T2I & 0.4842 & 2.5741 & 0.4868 & 2.7983 & 0.4470 & 2.5734 \\
VAR-512 & 0.1949 & 1.1867 & 0.1263 & 0.7472 & 0.1121 & 0.6281 \\
Infinity-d32 & 0.0790 & 0.4371 & 0.0098 & 0.0548 & 0.0026 & 0.0025 \\
Infinity-d64 & 0.0439 & 0.2978 & 0.0043 & 0.0237 & 0.0019 & 0.0000 \\
SD3.5L & 0.0002 & 0.0000 & 0.0004 & 0.0025 & 0.0045 & 0.0062 \\
FLUX.1-dev & 0.0521 & 0.4872 & 0.0004 & 0.0025 & 0.0013 & 0.0012 \\
GPT-4o & 0.0180 & 0.0836 & 0.0186 & 0.1034 & 0.0173 & 0.0784 \\
\midrule
\bottomrule
\end{tabular}
}
\label{tab:ocr_comparison}
\end{table}

The three OCR systems show consistent overall performance trends, with some differences in individual model rankings. High-performing VTs such as BSQ-VIT, Infinity-d64, and FLUX.1-dev consistently outperform weaker tokenizers such as Titok variants or MaskBiT. The results in Table~\ref{tab:ocr_comparison} support the main text-preservation findings across OCR systems.

\section{Experimental Settings and Environments}\label{apd:experimental_setting}

\textbf{Experimental Settings.}
We evaluate a range of SOTA VTs across the three core tasks in VTBench: Image Reconstruction, Detail Preservation, and Text Preservation. For each task, we use a fixed evaluation protocol to ensure consistency across models. All tokenizers are evaluated in an inference-only setting using publicly available pretrained weights. Where applicable, we follow each model's official preprocessing pipeline and input size specification (see Table~\ref{tbl:task1}). The FlowMo high-resolution and varying-resolution tests retain the whole input image and do not use patchwise inference. FlowMo Hi uses a fixed 14,336-bit latent budget: 0.219 nominal bits per pixel at its $256 \times 256$ training resolution and 0.0137 at $1024 \times 1024$. Reduced capacity per pixel and the training-resolution mismatch are possible contributors to corruption. This whole-image evaluation differs from FlowMo's patchwise high-resolution inference. For text preservation evaluation, OCR is applied to both original and reconstructed images before computing CER and WER. We use the same set of 50,000 ImageNet validation images and 100 samples per text benchmark (movie posters, abstracts, multilingual) for all models.

\textbf{Environments.}
All experiments are conducted using 4 NVIDIA A100 80GB GPUs with 251GB memory. Image preprocessing, reconstruction, and OCR evaluation are run on a standardized, reproducible pipeline. All results are collected on the same hardware and software stack to ensure fair comparisons.

\begin{figure*}[t]
  \centering
  \includegraphics[width=1\textwidth]{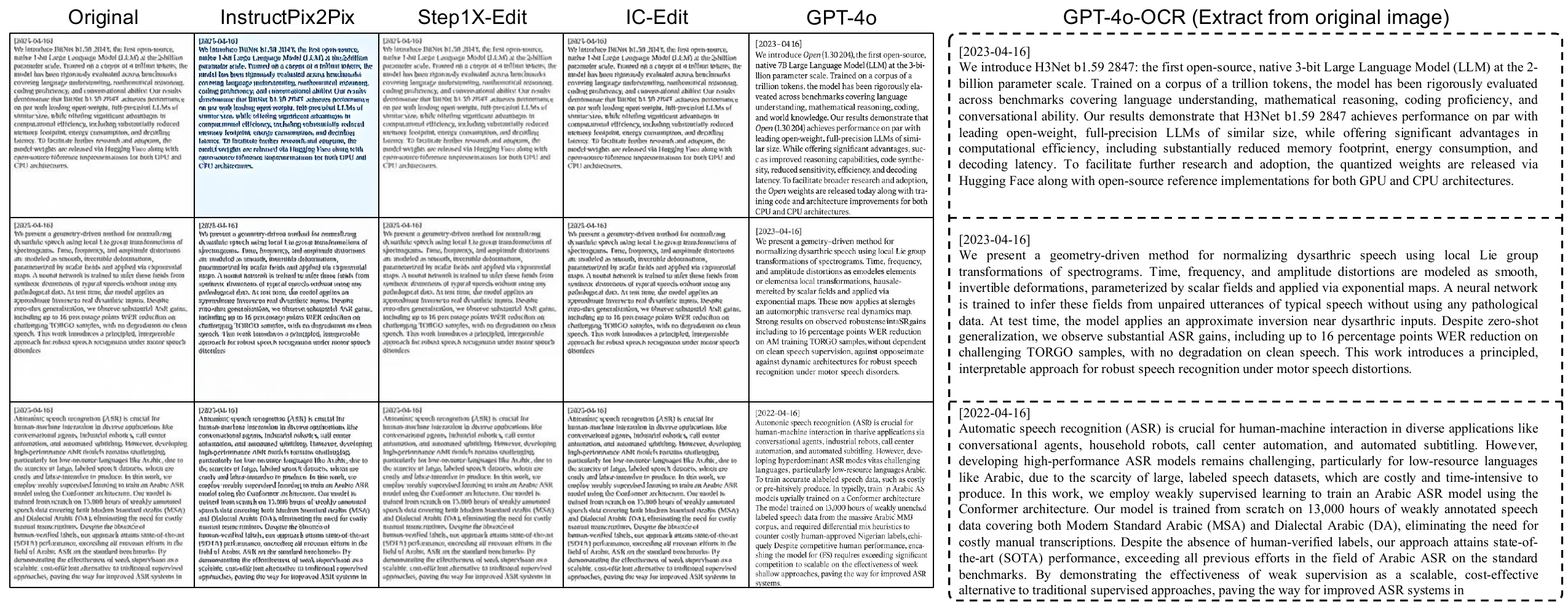}
  \caption{Comparison of image editing using the prompt \textit{``Please recreate the exact same image without any alterations.''} with the original image. The original images contain blurred text that is not recognizable to humans. The rightmost column displays the OCR-extracted text from the original image using GPT-4o using the prompt \textit{``Directly output what is written in the image.''}}
  \label{fig:gpt4o_comparison}
\end{figure*}

\section{Additional Experiments of GPT-4o Image Generation}\label{apd:gpt4o-image-generation}

To further understand GPT-4o's architecture and VT, we conducted a set of controlled experiments focused on its capabilities in font recognition and fine-grained text editing. These experiments examine GPT-4o's output behavior.

\subsection{Image Reconstruction from Low-Quality Inputs}
In this experiment, we provided GPT-4o with an intentionally degraded input image containing heavily distorted or unreadable text. Our prompt explicitly instructed the model to \emph{recreate the image exactly without any alteration}. Interestingly, unlike diffusion-based models, which tend to replicate the visual noise and illegibility, GPT-4o instead produced an image where the text had been cleanly restored into readable, coherent sentences, as shown in Figure~\ref{fig:gpt4o_comparison}.

This behavior strongly deviates from the expected literal replication and instead suggests that GPT-4o engaged in a form of semantic hallucination. We hypothesize that GPT-4o first attempts to understand the image content through an internal representation, and in doing so, ``fills in'' the missing or noisy parts based on learned linguistic priors. To test this hypothesis, we queried GPT-4o to extract text from the original distorted image. Surprisingly, the model returned well-formed and highly plausible text, even though no readable characters are actually present in the image. Moreover, the extracted text closely matched the regenerated image, indicating a strong coupling between image understanding and generation.

These observations suggest that GPT-4o has inherited a strong capacity for image understanding. The model does not simply replicate noisy input, but instead attempts to semantically interpret and  ``restore'' the image content, even when such interpretation involves hallucination. This behavior implies that GPT-4o's backbone is not a generic image generation model, but one that possesses robust multimodal reasoning capabilities. We hypothesize that this capability arises either from: (1) \textbf{A shared autoregressive backbone} pretrained or co-trained for both language and vision tasks, enabling semantic-level image understanding and generation, or (2) \textbf{A language-centric model} (e.g., GPT-4) that has been subsequently fine-tuned on multimodal tasks, thereby acquiring the ability to parse and reconstruct image content via linguistic priors.

In either case, the consistency between the model's interpretation and generation of text, despite degraded visual input, strongly supports the view that GPT-4o performs image generation not as a pixel-to-pixel transformation task, but as a language-informed semantic generation process.

\begin{figure*}[t]
  \centering
  \includegraphics[width=1\textwidth]{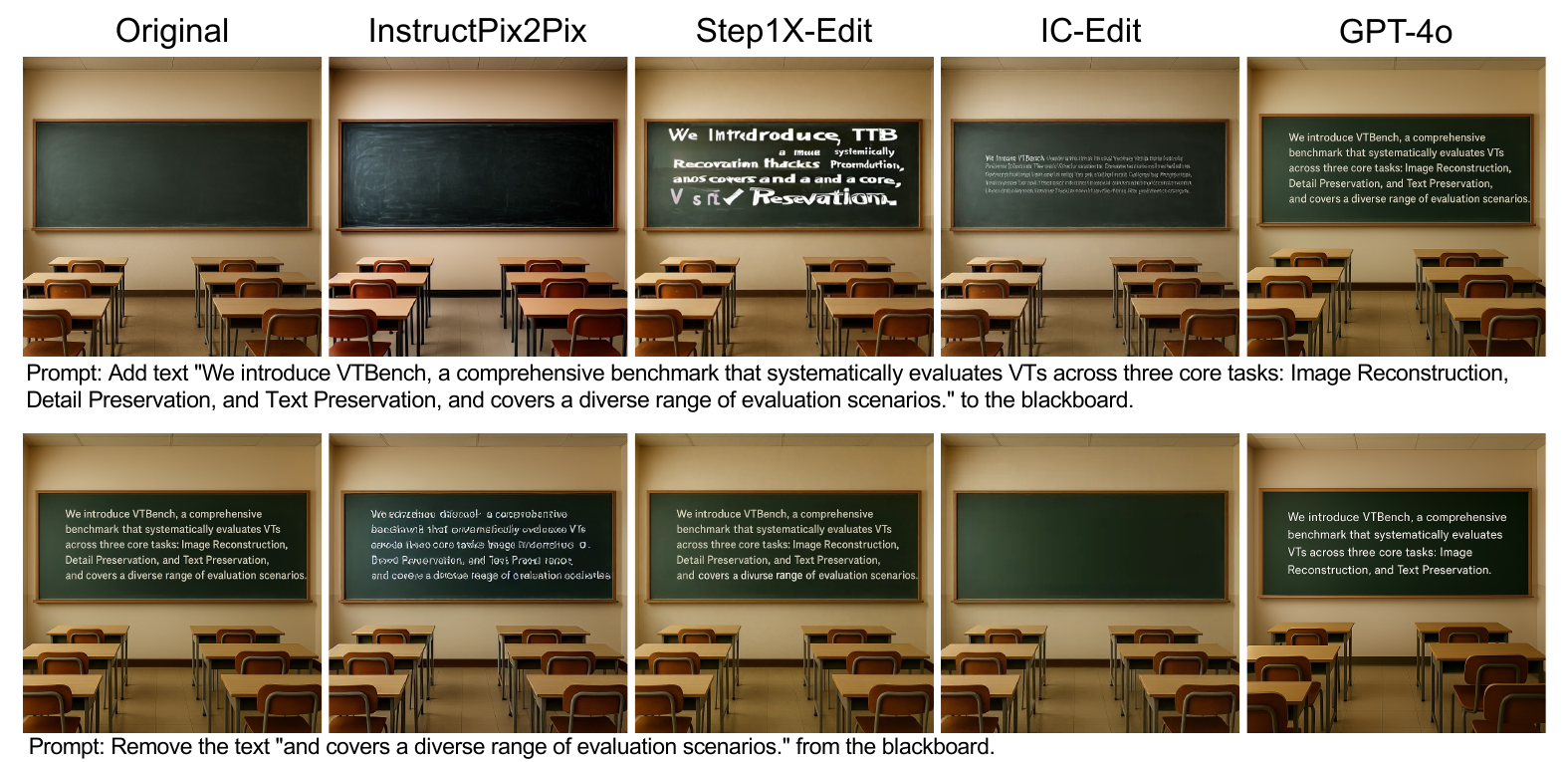}
  \vspace{-.2in}
  \caption{Text editing comparison across models, showing both text insertion (top) and removal (bottom) on a blackboard.}
  \label{fig:gpt4o_comparison_blackboard}
\end{figure*}

\subsection{Text Addition and Deletion in Scene Context}

To further explore GPT-4o's ability to understand and manipulate visual content, we conducted a series of experiments focused on localized text editing -- inserting and removing words from realistic visual scenes, such as blackboards and signage. These tasks require not only accurate prompt-following, but also semantic understanding of scene layout, font style, and background context.

In the text addition experiments, GPT-4o was prompted to insert specific phrases into designated regions of an image, as shown in Figure~\ref{fig:gpt4o_comparison_blackboard}. The results show remarkable spatial precision and stylistic coherence: the added text matched the surrounding content in font, size, alignment, and even lighting and shading. In the deletion experiments, GPT-4o is instructed to remove certain words. Rather than crudely masking out regions, the model naturally inpaints the background, reconstructing the texture and structure behind the deleted text with minimal artifacts.

These results highlight several properties of GPT-4o's image editing behavior:

\begin{itemize}[itemsep=2pt, topsep=0pt, leftmargin=20pt, parsep=0pt, partopsep=0pt]
    \item \textbf{Semantic-level image editing}: The model does not simply manipulate pixels, but appears to operate at a higher level of abstraction. It understands which parts of the image correspond to the textual prompt and modifies them while preserving global visual coherence.
    \item \textbf{Language-conditioned spatial reasoning}: The edits are guided by natural language, which suggests a strong alignment between textual and visual representations. This is indicative of a shared or tightly coupled multimodal representation space.
    \item \textbf{Unified generation and understanding backbone}: The precision and fidelity of the edits, especially the seamless integration of new content, suggest that the same underlying autoregressive model is responsible for both interpreting the image and generating the modified output. This unified design contrasts with traditional pipelines that separate perception and generation.
\end{itemize}

Together with the hallucination behavior observed in the font reconstruction experiment, these findings support the view that GPT-4o integrates image understanding and image generation into a single, coherent process. These capabilities motivate visual representations that preserve fine-grained appearance, semantic structure, and spatial context.

\begin{table*}[t]
\centering
\small
\caption{SBER-MoVQGAN results on Detail Preservation (100 images). Bold indicates the best result among the three configurations.}
\label{tab:sber_details}
\begin{tabular}{lcccc}
\toprule
SBER-MoVQGAN & PSNR$\uparrow$ & SSIM$\uparrow$ & LPIPS$\downarrow$ & FID$\downarrow$ \\
\midrule
67M & 19.7710 & 0.5505 & 0.0819 & 47.1139 \\
102M & 20.0695 & 0.5811 & 0.0739 & 39.2658 \\
270M & \textbf{20.3097} & \textbf{0.6009} & \textbf{0.0724} & \textbf{33.3104} \\
\bottomrule
\end{tabular}
\par\medskip
\caption{SBER-MoVQGAN results on Movie Posters (100 images). Bold indicates the best result among the three configurations.}
\label{tab:sber_posters}
\begin{tabular}{lcccccc}
\toprule
SBER-MoVQGAN & PSNR$\uparrow$ & SSIM$\uparrow$ & LPIPS$\downarrow$ & FID$\downarrow$ & CER$\downarrow$ & WER$\downarrow$ \\
\midrule
67M & 28.9414 & 0.8553 & 0.0272 & 33.0401 & 0.2281 & 1.2668 \\
102M & 29.8453 & 0.8637 & 0.0238 & 29.7922 & 0.1552 & 0.8619 \\
270M & \textbf{30.1748} & \textbf{0.8672} & \textbf{0.0228} & \textbf{27.9694} & \textbf{0.1169} & \textbf{0.6323} \\
\bottomrule
\end{tabular}
\par\medskip
\caption{SBER-MoVQGAN results on English ArXiv Abstracts (100 images). Bold indicates the best result among the three configurations.}
\label{tab:sber_abstracts}
\begin{tabular}{lcccccc}
\toprule
SBER-MoVQGAN & PSNR$\uparrow$ & SSIM$\uparrow$ & LPIPS$\downarrow$ & FID$\downarrow$ & CER$\downarrow$ & WER$\downarrow$ \\
\midrule
67M & 18.5920 & 0.8441 & 0.0384 & 5.4837 & 0.5089 & 3.7226 \\
102M & 19.9180 & 0.8867 & 0.0306 & \textbf{3.9158} & 0.2510 & 1.8385 \\
270M & \textbf{20.6366} & \textbf{0.9034} & \textbf{0.0276} & 4.2655 & \textbf{0.1773} & \textbf{1.2980} \\
\bottomrule
\end{tabular}
\end{table*}

\begin{figure*}[t]
  \centering
  \includegraphics[width=1\textwidth]{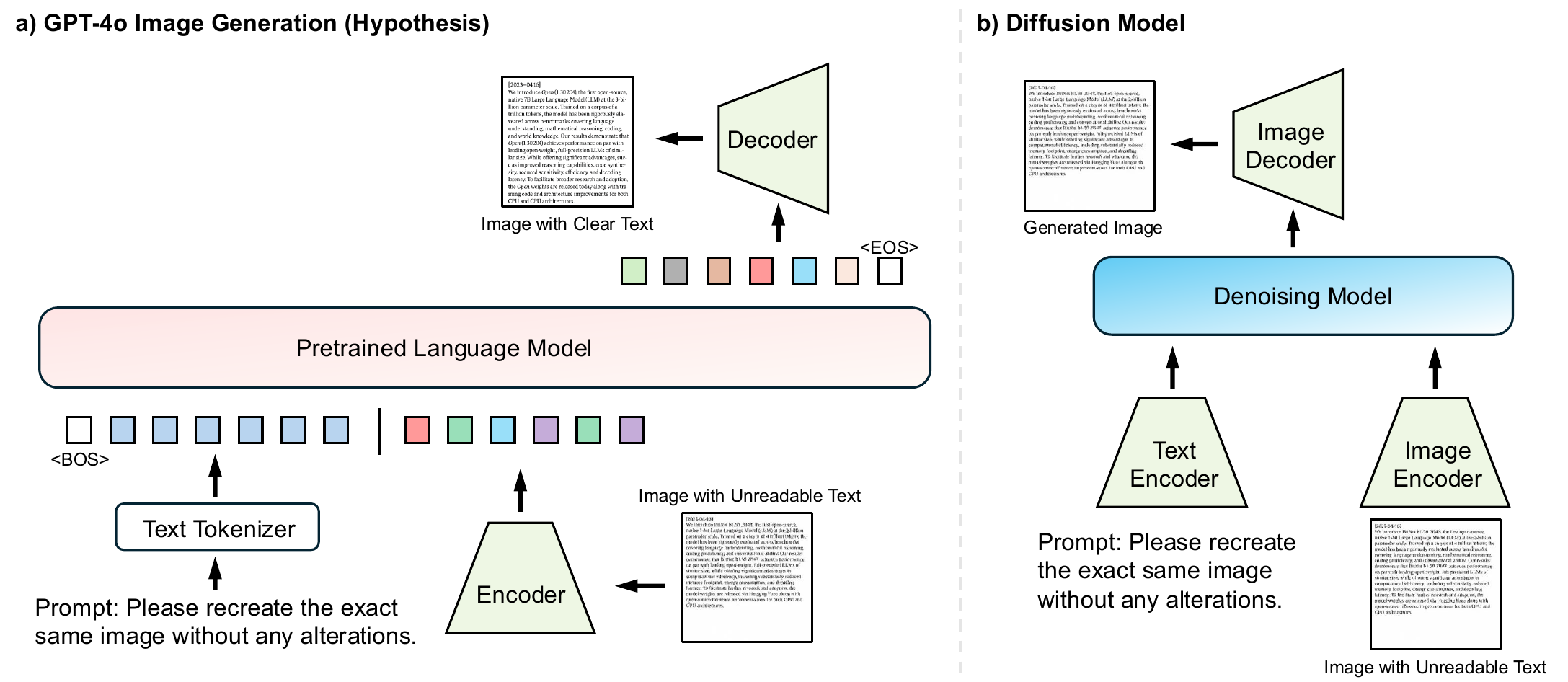}
  \caption{Comparison of two hypothesized architectures for image generation.}
  \label{fig:gpt4o_arch}
\end{figure*}

\subsection{Autoregressive vs. Diffusion: Implications for GPT-4o.}

As summarized in Figure~\ref{fig:gpt4o_arch}, we hypothesize that GPT-4o adopts a unified autoregressive architecture rather than a diffusion-based pipeline. This hypothesis is supported by the following observations:

\begin{itemize}[itemsep=2pt, topsep=0pt, leftmargin=20pt, parsep=0pt, partopsep=0pt]
    \item \textbf{Semantic hallucination through language priors:} GPT-4o's tendency to ``restore'' unreadable text into coherent language, despite prompts requesting exact replication suggests that its image understanding is conditioned by a language model prior. Such behavior is natural in an autoregressive framework where image and text tokens are processed in a shared sequence and jointly influenced by a pretrained language model.
    
    \item \textbf{Language-grounded editing and generation:} The model demonstrates precise text insertion and deletion aligned with natural language instructions, which requires not only generation capability but also deep semantic understanding of both visual context and instruction intent. These outputs show that image synthesis can respond to detailed linguistic instructions.

    \item \textbf{Modality reuse and efficiency:} Under the autoregressive hypothesis, a shared language-model backbone could provide common priors for textual and visual reasoning.

    \item \textbf{Failure mode contrast:} Diffusion models tend to replicate noise patterns or preserve unreadability when asked to reconstruct degraded text, faithfully following low-level visual patterns. GPT-4o, on the other hand, produces semantically enriched outputs, even when that contradicts pixel-level fidelity, indicative of language-centric generation.
\end{itemize}

\begin{figure*}[t]
  \centering
  \includegraphics[width=.83\textwidth]{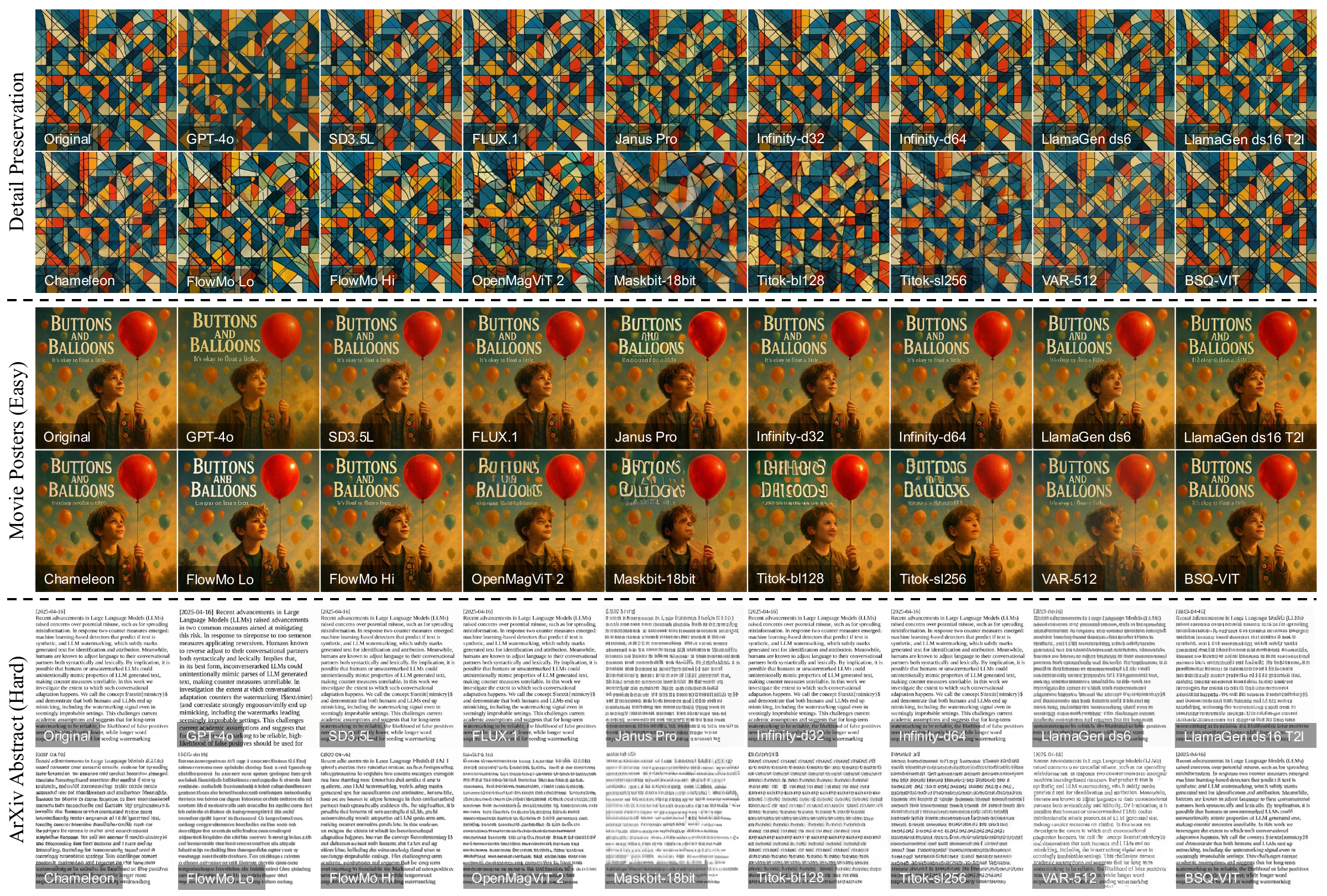}
  \caption{Qualitative results of detail preservation and text preservation.}
  \label{fig:task2_examples}
\end{figure*}

\begin{table}[htbp]
\centering
\caption{FID point estimates and bootstrap standard errors from 2,000 image-level resamples across three 100-image datasets.}
\label{tab:bootstrap_fid}
\resizebox{\columnwidth}{!}{%
\begin{tabular}{lccc}
\toprule
Method & High Resolution & Movie Posters & ArXiv Abstracts \\
\midrule
FlowMo Lo & $465.35 \pm 8.74$ & $55.07 \pm 2.42$ & $9.19 \pm 0.37$ \\
FlowMo Hi & $439.75 \pm 9.11$ & $40.70 \pm 1.85$ & $6.69 \pm 0.33$ \\
MaskBiT 16-bit & $24.25 \pm 1.42$ & $67.35 \pm 2.91$ & $27.99 \pm 0.88$ \\
MaskBiT 18-bit & $23.30 \pm 1.27$ & $64.08 \pm 2.56$ & $26.21 \pm 0.75$ \\
\bottomrule
\end{tabular}%
}
\end{table}

Together, these insights point toward GPT-4o employing an autoregressive backbone with a high-capacity visual tokenizer that supports semantic, instruction-driven generation. Under this hypothesis, a shared backbone could align image understanding and synthesis with language modeling.

\subsection{GPT-4o's Visual Tokenizer}
Despite the impressive capabilities exhibited by GPT-4o in both image understanding and generation, its VT remains proprietary and undisclosed. Nevertheless, its behavior provides strong indirect evidence of a highly expressive VT design: it preserves fine-grained textual structure, aligns visual and linguistic semantics, and supports high-resolution, editable image representations. Unlike many existing discrete tokenizers that struggle with text reconstruction and spatial fidelity, GPT-4o's VT appears to encode images in a way that is not only compact, but also semantically rich and compatible with autoregressive decoding.

To replicate such a tokenizer, future work must prioritize three core capabilities: 
\begin{itemize}[itemsep=2pt, topsep=0pt, leftmargin=20pt, parsep=0pt, partopsep=0pt]
\item \textbf{High-fidelity image reconstruction:} The tokenizer should retain sufficient visual detail to allow accurate reconstruction of the input image, ideally matching the performance of state-of-the-art continuous VAEs in terms of perceptual quality and structure preservation.
\item \textbf{Resolution scalability:} It should generalize across varying image sizes and support high-resolution inputs without introducing artifacts or requiring fixed input dimensions, a common limitation of many existing discrete tokenizers.
\item \textbf{Compatibility with downstream language models:} The output token sequences must be structurally and semantically aligned with the expectations of large language models (LLMs), enabling seamless integration into autoregressive multimodal frameworks for joint reasoning and generation.
\end{itemize}
Architectures such as residual next-scale VAEs, hierarchical quantization, or continuous tokenization may serve as promising directions. Given the lack of public access to GPT-4o's VT, benchmarks like VTBench offer a practical path forward: by systematically evaluating VT components in isolation, researchers can identify weaknesses, measure progress, and guide the development of open-source alternatives that approach or surpass GPT-4o's performance.

\section{Discussion}

\textbf{Limitation of Existing VTs.} Discrete VTs show lower overall reconstruction fidelity, detail preservation, and text accuracy than continuous VAEs, particularly under high-resolution, variable-size, and multilingual conditions. Many models are constrained by fixed input sizes and struggle to retain semantic structure in complex settings. Moreover, the absence of a strong, open-source VT limits progress in large-scale AR models. While LLMs continue to scale and improve, their performance is increasingly bottlenecked by the quality of visual tokenization. There is an urgent need for a resolution-flexible, semantically robust, and reusable VT that can keep pace with the capabilities of modern LLMs. We hope that VTBench can help accelerate research in this direction by providing a unified framework for diagnosis, comparison, and future development.

\textbf{Insights from GPT-4o Image Generation.} We present additional experiments comparing GPT-4o's image generation quality with that of diffusion models in Appendix~\ref{apd:gpt4o-image-generation}. The results suggest that GPT-4o may employ an autoregressive generation mechanism, aligning with prior research hypotheses~\citep{yan2025gpt}. GPT-4o appears to inherit the knowledge and reasoning capabilities of LLMs while achieving strong visual synthesis quality, indicating its potential as a unified multimodal model. Although the exact architecture of GPT-4o's VT remains undisclosed, the high quality of its generated images implies a highly capable VT design. Based on qualitative analysis, we hypothesize that GPT-4o may use a residual next-scale VAE (RVAE)~\citep{VAR, infinity} or a diffusion-based encoder-decoder~\citep{yan2025gpt}. We consider the former more likely, as residual tokenization naturally aligns with the input-output format of LLMs and supports AR generation within a unified framework. This structure would allow the model to seamlessly integrate image understanding and generation, facilitating effective multimodal learning.

\section{Additional Qualitative Results}\label{apd:additional_visualization}

\begin{figure*}[t]
  \centering
  \includegraphics[width=1\textwidth]{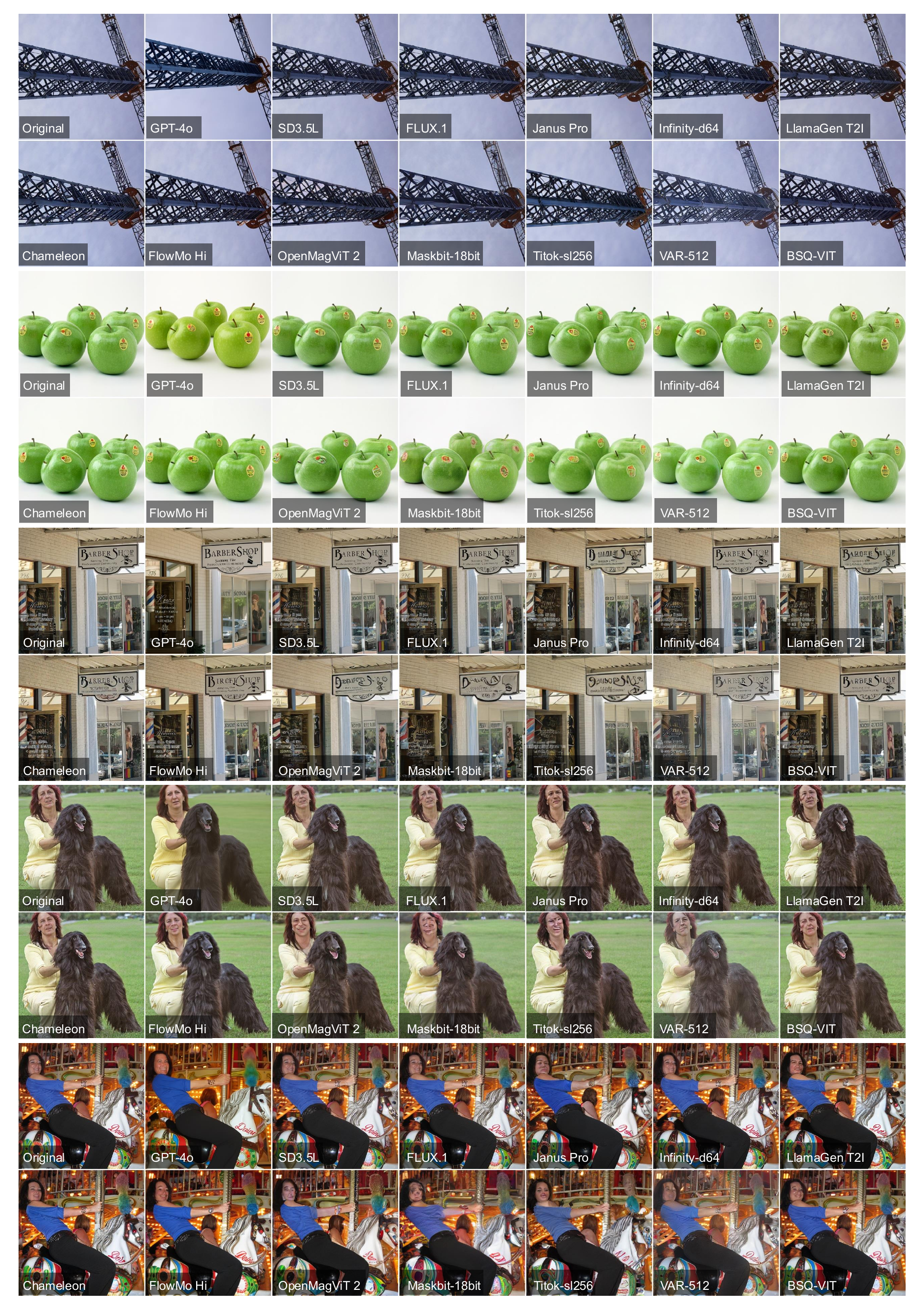}
  \caption{Additional qualitative results for ImageNet in Task 1.}
  \label{fig:additional_imagenet}
\end{figure*}

\begin{figure*}[t]
  \centering
  \includegraphics[width=1\textwidth]{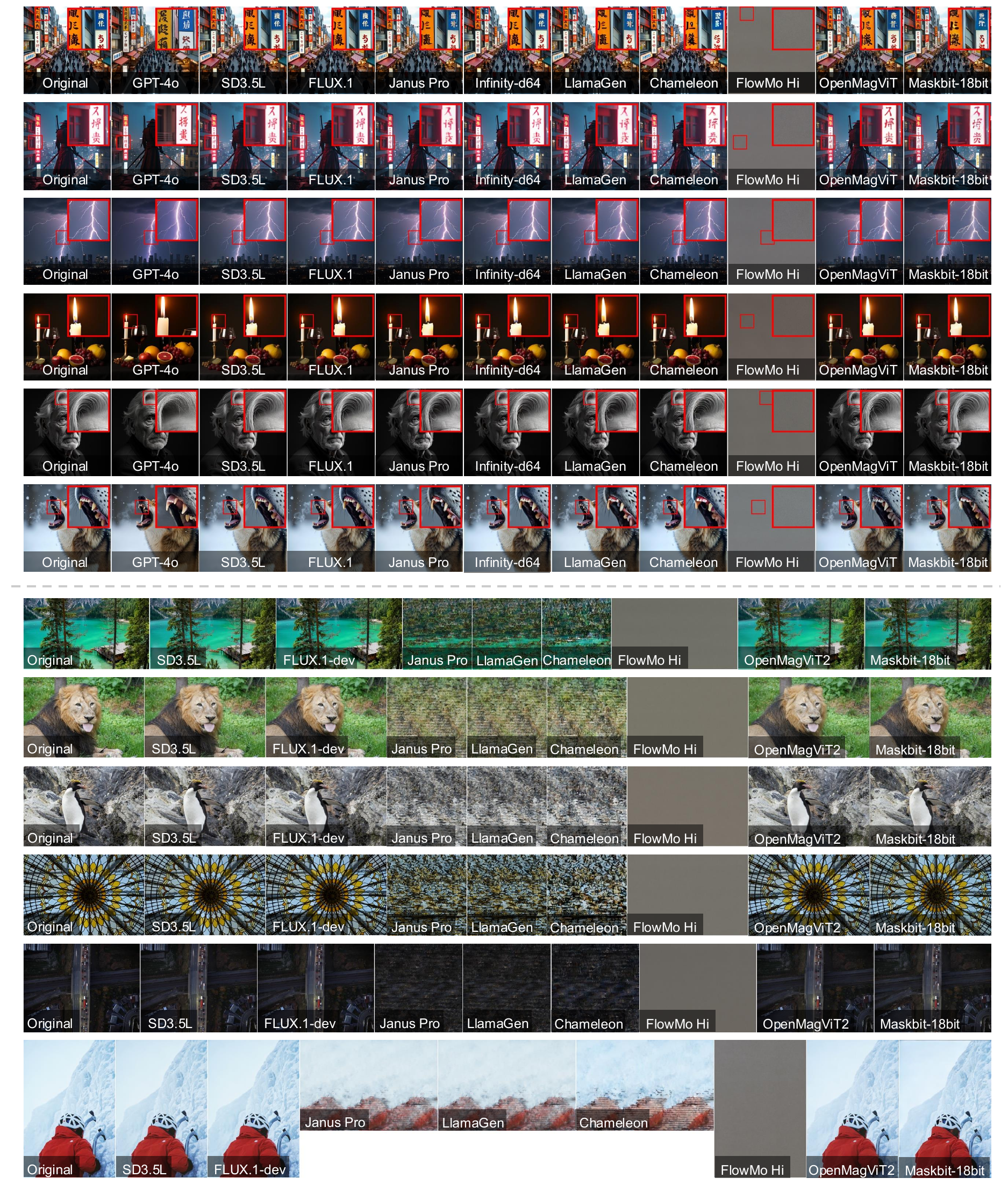}
  \caption{Visualized examples for high-resolution (top) and varying-resolution (bottom) of task 1.}
  \label{fig:additional_highresolution}
\end{figure*}

\begin{figure*}[t]
  \centering
  \includegraphics[width=1\textwidth]{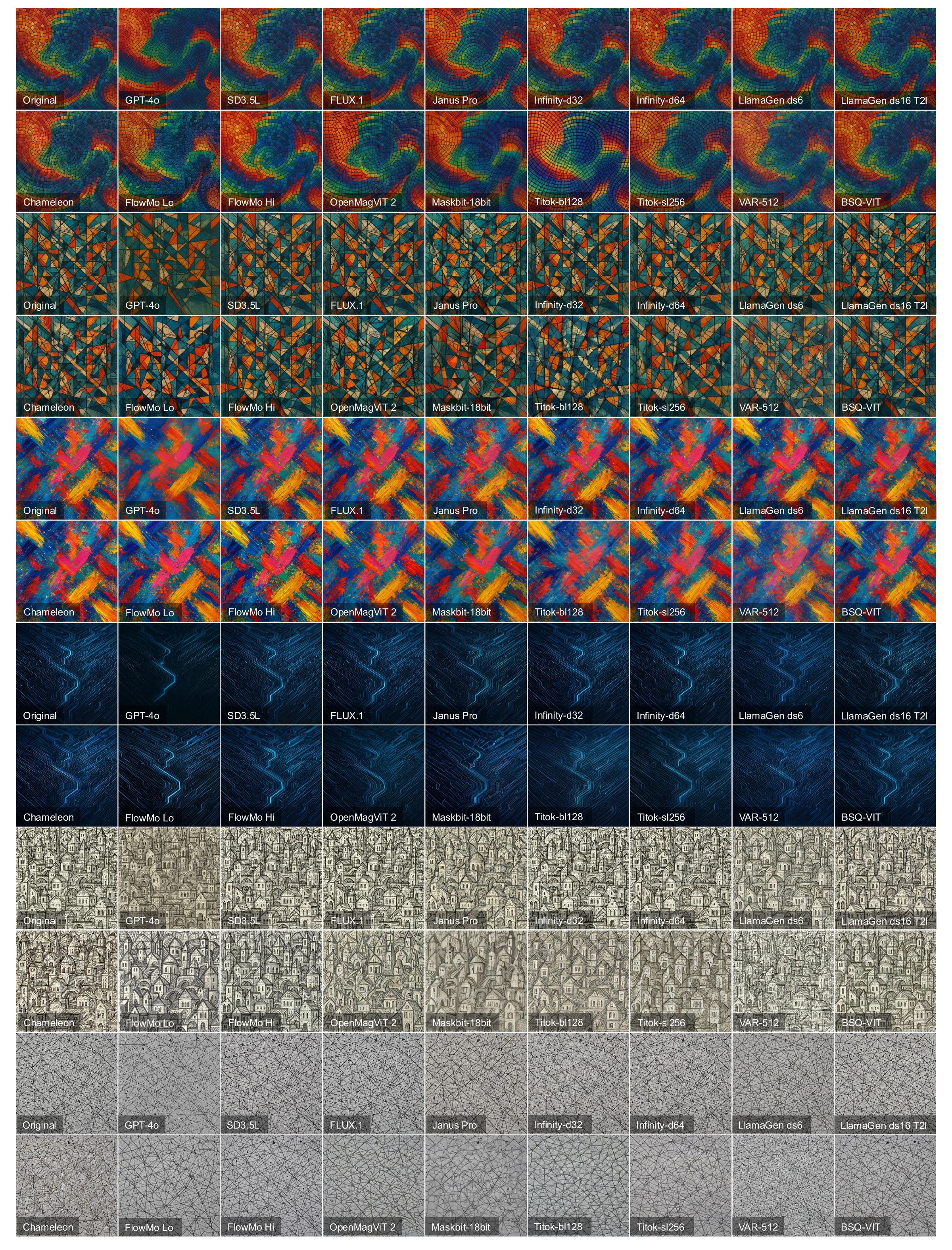}
  \caption{Additional qualitative results for detail preservation (Task 2).}
  \label{fig:additional_task2}
\end{figure*}

\begin{figure*}[t]
  \centering
  \includegraphics[width=1\textwidth]{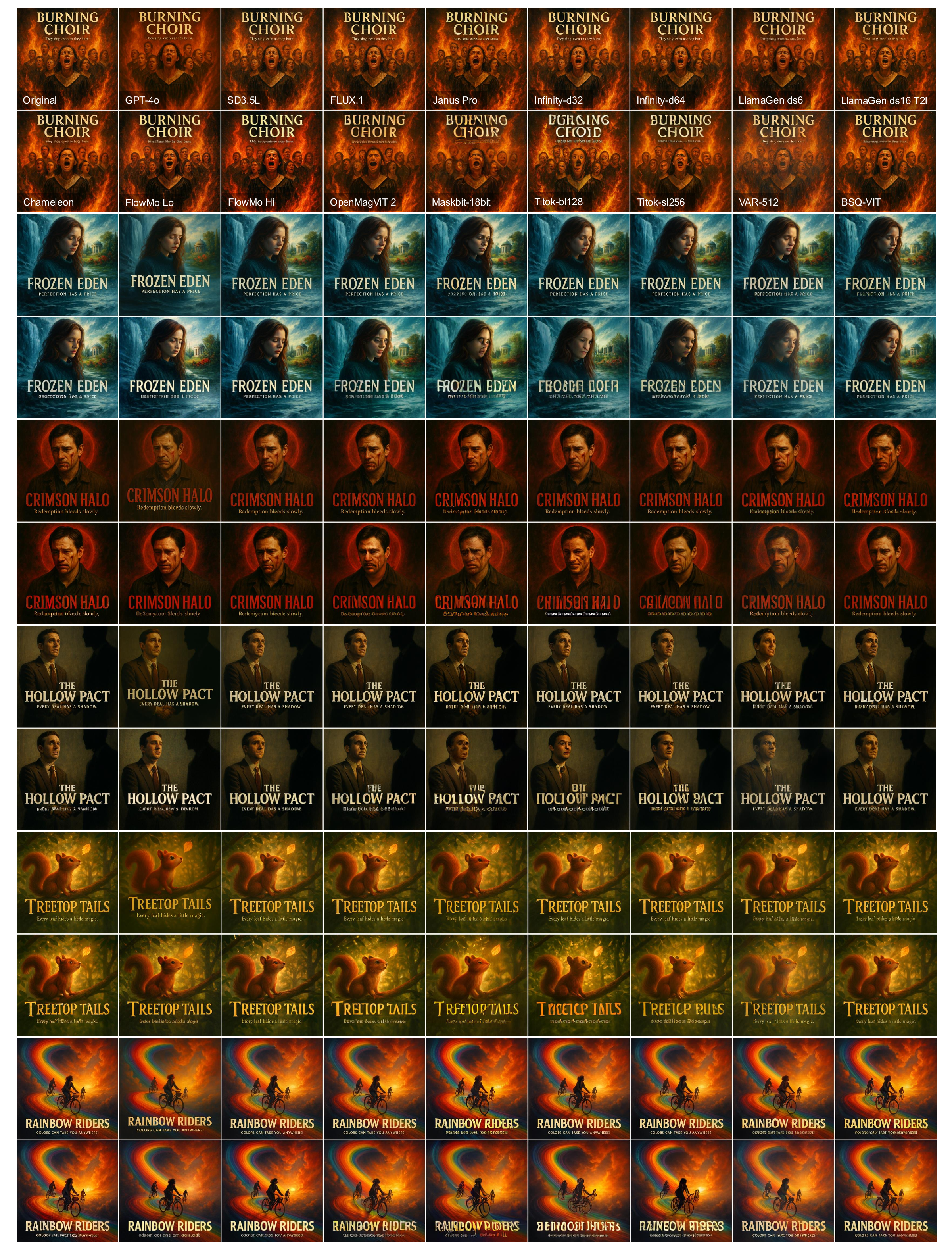}
  \caption{Additional qualitative results for movie posters in Task 3. The corresponding model names are listed in the first row.}
  \label{fig:additional_task3_movie_posters}
\end{figure*}

\begin{figure*}[t]
  \centering
  \includegraphics[width=1\textwidth]{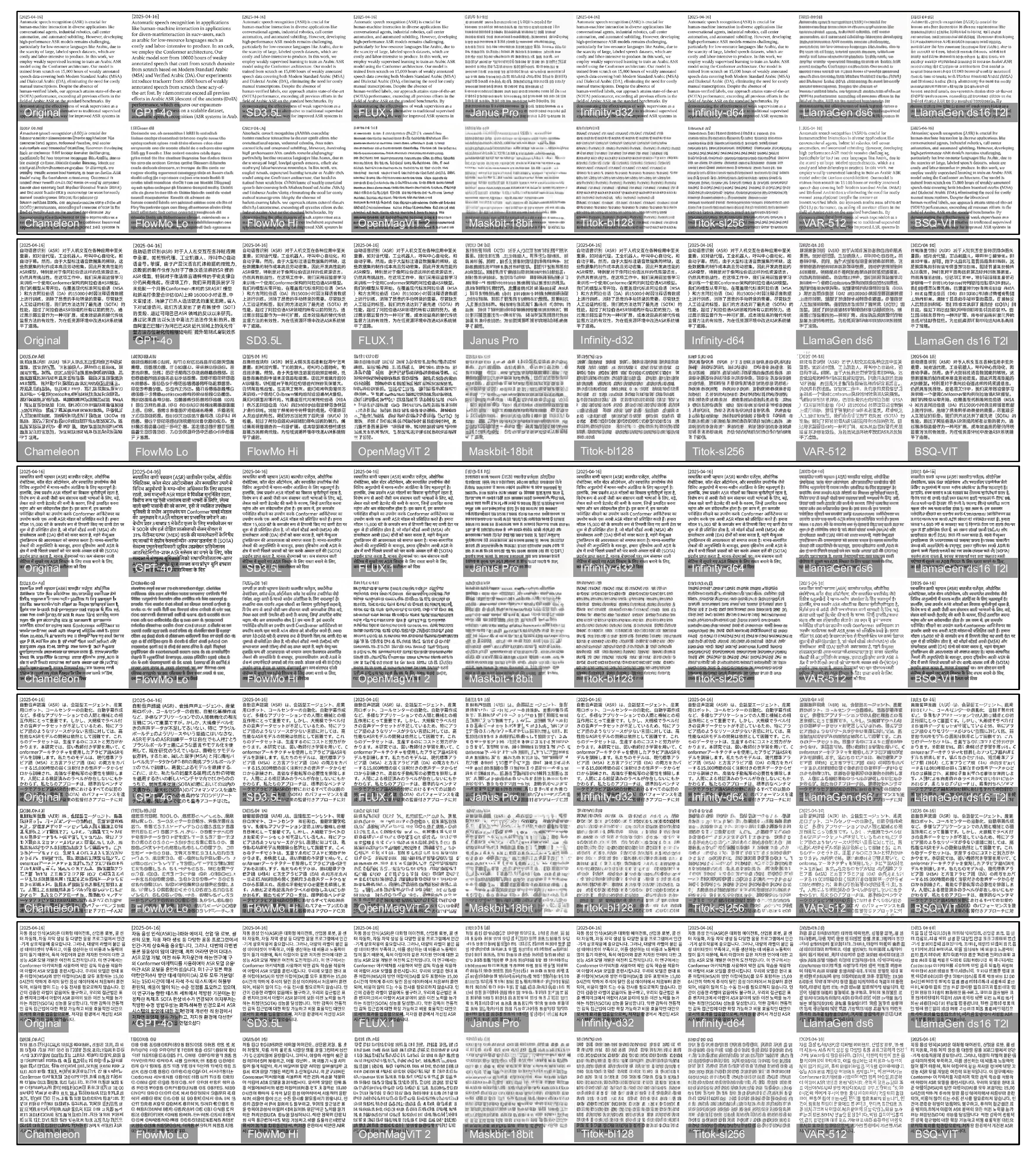}
  \caption{Additional qualitative results showing one ArXiv abstract and its corresponding multilingual versions (Chinese, Japanese, Korean, and Hindi) reconstructed by various visual tokenizers.}
  \label{fig:additional_task3_arxiv_multilingual}
\end{figure*}

To complement the quantitative evaluations presented in the main paper, we provide additional qualitative results that illustrate the reconstruction quality, text preservation, and detail retention of various visual tokenizers (Figure~\ref{fig:task2_examples} and Figure~\ref{fig:additional_imagenet} to \ref{fig:additional_task3_arxiv_multilingual}).

\section*{Ethics Statement}
All authors have read and will adhere to the ACL Code of Ethics. Our study uses publicly available datasets (ImageNet-1k val, DIV2K test) and synthetically generated images created via programmatic prompts; no human subjects or personal data are involved. For arXiv abstracts, we reference public identifiers and provide rendering code rather than redistributing restricted text. We took care to avoid harmful or offensive content in the synthetic sets and to document fonts and layouts for multilingual renders to mitigate script-specific bias. Third-party models and assets are used under their respective licenses and appropriately credited. There are no conflicts of interest or external sponsorship influencing this work. We will release the dataset variants produced for VTBench and the full codebase to support transparency and community scrutiny.

\section*{Reproducibility Statement}
We provide an anonymized repository with complete evaluation code, configuration files, and instructions to reproduce all figures and tables. The repository includes exact preprocessing, metric implementations (PSNR, SSIM, LPIPS, FID, CER, WER), fixed splits for each task, prompts and rendering pipelines for synthetic data, and pointers or scripts to obtain any licensed sources. Software and hardware details and versioning are documented, and random seeds are fixed where applicable; sources of nondeterminism (e.g., external image APIs) are noted with our selection protocol. Upon publication, we will open-source both the VTBench dataset and the code to facilitate independent verification and extension.

\section{Bootstrap Uncertainty}\label{apd:bootstrap_uncertainty}
Table~\ref{tab:bootstrap_fid} reports FID point estimates with bootstrap standard errors from 2,000 image-level resamples for FlowMo Lo/Hi and MaskBiT 16/18-bit on High Resolution, Movie Posters, and English ArXiv Abstracts. Each dataset contains 100 images. Point estimates are rounded to two decimals here; Tables~\ref{tbl:task1} and~\ref{tbl:task3} retain their original precision.

\section{Targeted SBER-MoVQGAN Evaluation}\label{apd:sber_movqgan}
We evaluate the three released SBER-MoVQGAN configurations, 67M, 102M, and 270M, on the same 100-image Detail Preservation, Movie Posters, and English ArXiv Abstracts subsets used in the corresponding main-paper tasks, using the same evaluation protocol. Tables~\ref{tab:sber_details}--\ref{tab:sber_abstracts} report the results in the same metric order as the main tables.

\end{document}